\documentclass{article}

\usepackage[nonatbib, final]{neurips_2019}

\usepackage[utf8]{inputenc} 
\usepackage[T1]{fontenc}    
\usepackage{hyperref}       
\usepackage{url}            
\usepackage{booktabs}       
\usepackage{amsfonts}       
\usepackage{nicefrac}       
\usepackage{microtype}      
\usepackage{multirow}

\usepackage{listings}
\usepackage[ruled, vlined, linesnumbered]{algorithm2e}
\usepackage{wrapfig}
\usepackage{graphicx}
\usepackage{amsmath}
\usepackage{caption}
\usepackage{subcaption}
\usepackage{blindtext}

\usepackage[x11names]{xcolor}
\usepackage{paralist}
\usepackage{enumitem}

\newcommand{\breakcell}[1]{\begin{tabular}{@{}l} #1 \end{tabular}}
\usepackage[disable]{todonotes}

\newcommand{\code}[1]{\texttt{#1}}
\lstdefinestyle{Python}{
    basicstyle      = \ttfamily \lst@ifdisplaystyle\tiny\fi,
    keywordstyle    = \color{RoyalBlue3},
    stringstyle     = \color{OrangeRed3},
    commentstyle    = \color{Wheat3}\ttfamily,
    frame=single,
    framerule=0.8pt,
    numbers=right,
    numbersep=0.2cm
}
\title{Towards modular and programmable architecture search}

\author{%
  Renato Negrinho$^{1}$ \thanks{Part of this work was done while the first author was a research scientist at Petuum.} \qquad Darshan Patil$^{1}$ \qquad Nghia Le$^{1}$ \qquad Daniel Ferreira$^{2}$\\
    \textbf{Matthew Gormley}$^{1}$ \qquad \textbf{Geoffrey Gordon}$^{1,3}$
    \\
Carnegie Mellon University$^{1}$, TU Wien$^{2}$, Microsoft Research Montreal$^{3}$\\
}

\begin{document}

\maketitle

\begin{abstract}
    Neural architecture search methods are able to find high performance deep learning architectures with minimal effort from an expert~\cite{elsken2018neural}.
    However, current systems focus on specific use-cases (e.g. convolutional image classifiers and recurrent language models), making them unsuitable for general use-cases that an expert might wish to write.
    Hyperparameter optimization systems~\cite{bergstra2013hyperopt,snoek2012practical,falkner2018bohb} are general-purpose but lack the constructs needed for easy application to architecture search.
    In this work, we propose a formal language for encoding search spaces over general computational graphs.
    The language constructs allow us to write modular, composable, and reusable search space encodings and to reason about search space design.
    We use our language to encode search spaces from the architecture search literature.
    The language allows us to decouple the implementations of the search space and the search algorithm, allowing us to expose search spaces to search algorithms through a consistent interface.
    Our experiments show the ease with which we can experiment with different combinations of search spaces and search algorithms without having to implement each combination from scratch.
    We release an implementation of our language with this paper\footnote{Visit \url{https://github.com/negrinho/deep_architect} for code and documentation.}.
\end{abstract}

\section{Introduction}
\label{introduction}

Architecture search has the potential to transform machine learning workflows.
High performance deep learning architectures are often manually designed through a trial-and-error process that amounts to trying slight variations of known high performance architectures.
Recently, architecture search techniques have shown tremendous potential by improving on handcrafted architectures, both by improving state-of-the-art performance and by finding better tradeoffs between computation and performance.
Unfortunately, current systems fall short of providing strong support for general architecture search use-cases.

Hyperparameter optimization systems~\cite{bergstra2013hyperopt,snoek2012practical,falkner2018bohb,smac-2017} are not designed specifically for architecture search use-cases and therefore do not introduce constructs that allow experts to implement these use-cases efficiently, e.g., easily writing new search spaces over architectures.
Using hyperparameter optimization systems for an architecture search use-case requires the expert to write the encoding for the search space over architectures as a conditional hyperparameter space and to write the mapping from hyperparameter values to the architecture to be evaluated.
Hyperparameter optimization systems are completely agnostic that their hyperparameter spaces encode search spaces over architectures.

By contrast, architecture search systems~\cite{elsken2018neural} are in their infancy, being tied to specific use-cases (e.g., either reproducing results reported in a paper or concrete systems, e.g., for searching over Scikit-Learn pipelines~\cite{olson2016tpot}) and therefore lack support for general architecture search workflows.
For example, current implementations of architecture search methods rely on ad-hoc encodings for search spaces, providing limited extensibility and programmability for new work to build on.
For example, implementations of the search space and search algorithm are often intertwined, requiring substantial coding effort to try new search spaces or search algorithms.

\paragraph{Contributions}
We describe a modular language for encoding search spaces over general computational graphs.
We aim to improve the programmability, modularity, and reusability of architecture search systems.
We are able to use the language constructs to encode search spaces in the literature.
Furthermore, these constructs allow the expert to create new search spaces and modify existing ones in structured ways.
Search spaces expressed in the language are exposed to search algorithms under a consistent interface, decoupling the implementations of search spaces and search algorithms.
We showcase these functionalities by easily comparing search spaces and search algorithms from the architecture search literature.
These properties will enable better architecture search research by making it easier to benchmark and reuse search algorithms and search spaces.

\section{Related work}
\label{sec:related_work}

\paragraph{Hyperparameter optimization}
Algorithms for hyperparameter optimization often focus on small or simple hyperparameter spaces (e.g., closed subsets of Euclidean space in low dimensions).
Hyperparameters might be categorical (e.g., choice of regularizer) or continuous (e.g., learning rate and regularization constant).
Gaussian process Bayesian optimization~\cite{shahriari2016taking} and sequential model based optimization~\cite{hutter2011sequential} are two popular approaches.
Random search has been found to be competitive for hyperparameter optimization~\cite{bergstra2012random, li2017hyperband}.
Conditional hyperparameter spaces (i.e., where some hyperparameters may be available only for specific values of other hyperparameters) have also been considered~\cite{bergstra2011algorithms, bergstra2013making}.
Hyperparameter optimization systems (e.g. Hyperopt~\cite{bergstra2013hyperopt}, Spearmint~\cite{snoek2012practical}, SMAC~\cite{smac-2017, hutter2011sequential} and BOHB~\cite{falkner2018bohb}) are general-purpose and domain-independent.
Yet, they rely on the expert to distill the problem into an hyperparameter space and write the mapping from hyperparameter values to implementations.

\paragraph{Architecture search}
Contributions to architecture search often come in the form of search algorithms, evaluation strategies, and search spaces.
Researchers have considered a variety of search algorithms, including reinforcement learning~\cite{zoph2016neural}, evolutionary algorithms~\cite{real2018regularized, liu2017hierarchical}, MCTS~\cite{negrinho2017deeparchitect}, SMBO ~\cite{negrinho2017deeparchitect,liu2018progressive}, and Bayesian optimization~\cite{kandasamy2018neural}.
Most search spaces have been proposed for recurrent or convolutional architectures~\cite{zoph2016neural,real2018regularized, liu2017hierarchical} focusing on image classification (CIFAR-10) and language modeling (PTB).
Architecture search encodes much of the architecture design in the search space (e.g., the connectivity structure of the computational graph, how many operations to use, their type, and values for specifying each operation chosen).
However, the literature has yet to provide a consistent method for designing and encoding such search spaces.
Systems such as Auto-Sklearn~\cite{feurer2015efficient}, TPOT~\cite{olson2016automating}, and Auto-Keras~\cite{jin2018efficient} have been developed for specific use-cases (e.g., Auto-Sklearn and TPOT focus on classification and regression of featurized vector data, Auto-Keras focus on image classification) and therefore support relatively rigid workflows.
The lack of focus on extensibility and programmability makes these systems unsuitable as frameworks for general architecture search research.

\section{Proposed approach: modular and programmable search spaces}
\label{sec:proposed_approach}

To maximize the impact of architecture search research, it is \emph{fundamental} to improve the programmability of architecture search tools\footnote{\emph{cf.} the effect of highly programmable deep learning frameworks on deep learning research and practice.}.
We move towards this goal by designing \emph{a language to write search spaces over computational graphs}.
We identify the following advantages for our language and search spaces encoded in it:
\begin{itemize}[leftmargin=2em]
\item \textbf{Similarity to computational graphs:}
    Writing a search space in our language is similar to writing a fixed computational graph in an existing deep learning framework.
    The main difference is that nodes in the graph may be search spaces rather than fixed operations (e.g., see Figure~\ref{fig:graph_transitions}).
    A search space maps to a single computational graph once all its hyperparameters have been assigned values (e.g., in frame \code{d} in Figure~\ref{fig:graph_transitions}).

\item \textbf{Modularity and reusability:}
    The building blocks of our search spaces are modules and hyperparameters.
    Search spaces are created through the composition of modules and their interactions.
    Implementing a new module only requires dealing with aspects local to the module.
    Modules and hyperparameters can be reused across search spaces, and new search spaces can be written by combining existing search spaces.
    Furthermore, our language supports search spaces in general domains (e.g., deep learning architectures or Scikit-Learn~\cite{pedregosa2011scikit} pipelines).

\item \textbf{Laziness:}
    A \emph{substitution module} delays the creation of a subsearch space until all hyperparameters of the substitution module are assigned values.
    Experts can use substitution modules to encode natural and complex conditional constructions by concerning themselves only with the conditional branch that is chosen.
    This is simpler than the support for conditional hyperparameter spaces provided by hyperparameter optimization tools, e.g., in  Hyperopt~\cite{bergstra2013hyperopt}, where all conditional branches need to be written down explicitly.
    Our language allows conditional constructs to be expressed implicitly through composition of language constructs (e.g., nesting substitution modules).
    Laziness also allows us to encode search spaces that can expand infinitely, which is not possible with current hyperparameter optimization tools (see Appendix~\ref{sec:infinite_search_spaces}).

\item \textbf{Automatic compilation to runnable computational graphs:}
    Once all choices in the search space are made, the single architecture corresponding to the terminal search space can be mapped to a runnable computational graph (see Algorithm~\ref{alg:graph_propagation}).
    By contrast, for general hyperparameter optimization tools this mapping has to be written manually by the expert.
\end{itemize}

\section{Components of the search space specification language}
\label{sec:language_components}

A search space is a graph (see Figure~\ref{fig:graph_transitions}) consisting of hyperparameters (either of type independent or dependent) and modules (either of type basic or substitution).
This section describes our language components and show encodings of simple search spaces in our Python implementation.
Figure~\ref{fig:graph_transitions} and the corresponding search space encoding in Figure~\ref{fig:example} are used as running examples.
Appendix~\ref{app:components} and Appendix~\ref{app:search_space_example} provide additional details and examples, e.g. the recurrent cell search space of~\cite{pham2018efficient}.

\paragraph{Independent hyperparameters}
The value of an independent hyperparameter is chosen from its set of possible values.
An independent hyperparameter is created with a set of possible values, but without a value assigned to it.
Exposing search spaces to search algorithms relies mainly on iteration over and value assignment to independent hyperparameters.
An independent hyperparameter in our implementation is instantiated as, for example, $\code{D([1, 2, 4, 8])}$.
In Figure~\ref{fig:graph_transitions}, \code{IH-1} has set of possible values $\{64, 128\}$ and is eventually assigned value $64$ (shown in frame \code{d}).

\paragraph{Dependent hyperparameters}
The value of a dependent hyperparameter is computed as a function of the values of the hyperparameters it depends on (see line 7 of Algorithm~\ref{alg:graph_transition}).
Dependent hyperparameters are useful to encode relations between hyperparameters, e.g., in a convolutional network search space, we may want the number of filters to increase after each spatial reduction.
In our implementation, a dependent hyperparameter is instantiated as, for example, \code{h = DependentHyperparameter(lambda dh: 2*dh["units"], \{"units": h\_units\})}.
In Figure~\ref{fig:graph_transitions}, in the transition from frame \code{a} to frame \code{b}, \code{IH-3} is assigned value 1, triggering the value assignment of \code{DH-1} according to its function \code{fn:2*x}.

\paragraph{Basic modules}

\begin{wrapfigure}{r}{0.5\textwidth}
\begin{lstlisting}[style = Python]
def one_layer_net():
    a_in, a_out = dropout(D([0.25, 0.5]))
    b_in, b_out = dense(D([100, 200, 300]))
    c_in, c_out = relu()
    a_out["out"].connect(b_in["in"])
    b_out["out"].connect(c_in["in"])
    return a_in, c_out
\end{lstlisting}
\caption{Search space over feedforward networks with dropout rate of $0.25$ or $0.5$, ReLU activations, and one hidden layer with $100$, $200$, or $300$ units.}
\label{fig:basic_only_search_space}
\end{wrapfigure}

A basic module implements computation that depends on the values of its properties.
Search spaces involving only basic modules and hyperparameters do not create new modules or hyperparameters, and therefore are fixed computational graphs (e.g., see frames \code{c} and \code{d} in Figure~\ref{fig:graph_transitions}).
Upon compilation, a basic module consumes the values of its inputs, performs computation, and publishes the results to its outputs (see Algorithm~\ref{alg:graph_propagation}).
Deep learning layers can be wrapped as basic modules, e.g., a fully connected layer can be wrapped as a single-input single-output basic module with one hyperparameter for the number of units.
In the search space in Figure~\ref{fig:basic_only_search_space}, \code{dropout}, \code{dense}, and \code{relu} are basic modules.
In Figure~\ref{fig:graph_transitions}, both frames \code{c} and \code{d} are search spaces with only basic modules and hyperparameters.
In the search space of frame \code{d}, all hyperparameters have been assigned values, and therefore the single architecture can be mapped to its implementation (e.g., in Tensorflow).

\paragraph{Substitution modules}

\begin{wrapfigure}{r}{0.5\textwidth}
\begin{lstlisting}[style = Python]
def multi_layer_net():
    h_or = D([0, 1])
    h_repeat = D([1, 2, 4])
    return siso_repeat(
        lambda: siso_sequential([
            dense(D([300])),
            siso_or([relu, tanh], h_or)
        ]), h_repeat)
\end{lstlisting}
\caption{Search space over feedforward networks with 1, 2, or 4 hidden layers and ReLU or tanh activations.}
\label{fig:substitution_search_space}
\end{wrapfigure}

Substitution modules encode structural transformations of the computational graph that are delayed\footnote{Substitution modules are inspired by delayed evaluation in programming languages.} until their hyperparameters are assigned values.
Similarly to a basic module, a substitution module has hyperparameters, inputs, and outputs.
Contrary to a basic module, a substitution module does not implement computation---it is substituted by a subsearch space (which depends on the values of its hyperparameters and may contain new substitution modules).
Substitution is triggered once all its hyperparameters have been assigned values.
Upon substitution, the module is removed from the search space and its connections are rerouted to the corresponding inputs and outputs of the generated subsearch space (see Algorithm~\ref{alg:graph_transition} for how substitutions are resolved).
For example, in the transition from frame \code{b} to frame \code{c} of Figure~\ref{fig:graph_transitions}, \code{IH-2} was assigned the value $1$ and \code{Dropout-1} and \code{IH-7} were created by the substitution of \code{Optional-1}.
The connections of \code{Optional-1} were rerouted to \code{Dropout-1}.
If \code{IH-2} had been assigned the value $0$, \code{Optional-1} would have been substituted by an identity basic module and no new hyperparameters would have been created.
Figure~\ref{fig:substitution_search_space} shows a search space using two substitution modules:
\code{siso\_or} chooses between \code{relu} and \code{tanh};
\code{siso\_repeat} chooses how many layers to include.
\code{siso\_sequential} is used to avoid multiple calls to \code{connect} as in Figure~\ref{fig:basic_only_search_space}.

\paragraph{Auxiliary functions}

\begin{wrapfigure}{r}{0.5\textwidth}
\begin{lstlisting}[style = Python]
def rnn_cell(hidden_fn, output_fn):
    h_inputs, h_outputs = hidden_fn()
    y_inputs, y_outputs = output_fn()
    h_outputs["out"].connect(y_inputs["in"])
    return h_inputs, y_outputs
\end{lstlisting}
    \caption{Auxiliary function to create the search space for the recurrent cell given functions that create the subsearch spaces.}
    \label{fig:simple_rnn}
\end{wrapfigure}

Auxiliary functions, while not components per se, help create complex search spaces.
Auxiliary functions might take functions that create search spaces and put them together into a larger search space.
For example, the search space in Figure~\ref{fig:simple_rnn} defines an auxiliary RNN cell that captures the high-level functional dependency: $h_t = q_h(x_t, h_{t-1})$ and $y_t = q_y(h_t)$.
We can instantiate a specific search space as \code{rnn\_cell(lambda: siso\_sequential([concat(2), one\_layer\_net()]), multi\_layer\_net)}.

\section{Example search space}
\label{sec:example}

\begin{wrapfigure}[16]{r}{0.5\textwidth}
\begin{lstlisting}[style = Python]
def search_space():
    h_n = D([1, 2, 4])
    h_ndep = DependentHyperparameter(
        lambda dh: 2 * dh["x"], {"x": h_n})

    c_inputs, c_outputs = conv2d(D([64, 128]))
    o_inputs, o_outputs = siso_optional(
        lambda: dropout(D([0.25, 0.5])), D([0, 1]))
    fn = lambda: conv2d(D([64, 128]))
    r1_inputs, r1_outputs = siso_repeat(fn, h_n)
    r2_inputs, r2_outputs = siso_repeat(fn, h_ndep)
    cc_inputs, cc_outputs = concat(2)

    o_inputs["in"].connect(c_outputs["out"])
    r1_inputs["in"].connect(o_outputs["out"])
    r2_inputs["in"].connect(o_outputs["out"])
    cc_inputs["in0"].connect(r1_outputs["out"])
    cc_inputs["in1"].connect(r2_outputs["out"])
    return c_inputs, cc_outputs
\end{lstlisting}
\caption{
Simple search space showcasing all language components.
See also Figure~\ref{fig:graph_transitions}.}
\label{fig:example}
\end{wrapfigure}

\begin{figure}[tbp]
    \centering
\includegraphics[width=0.85\textwidth]{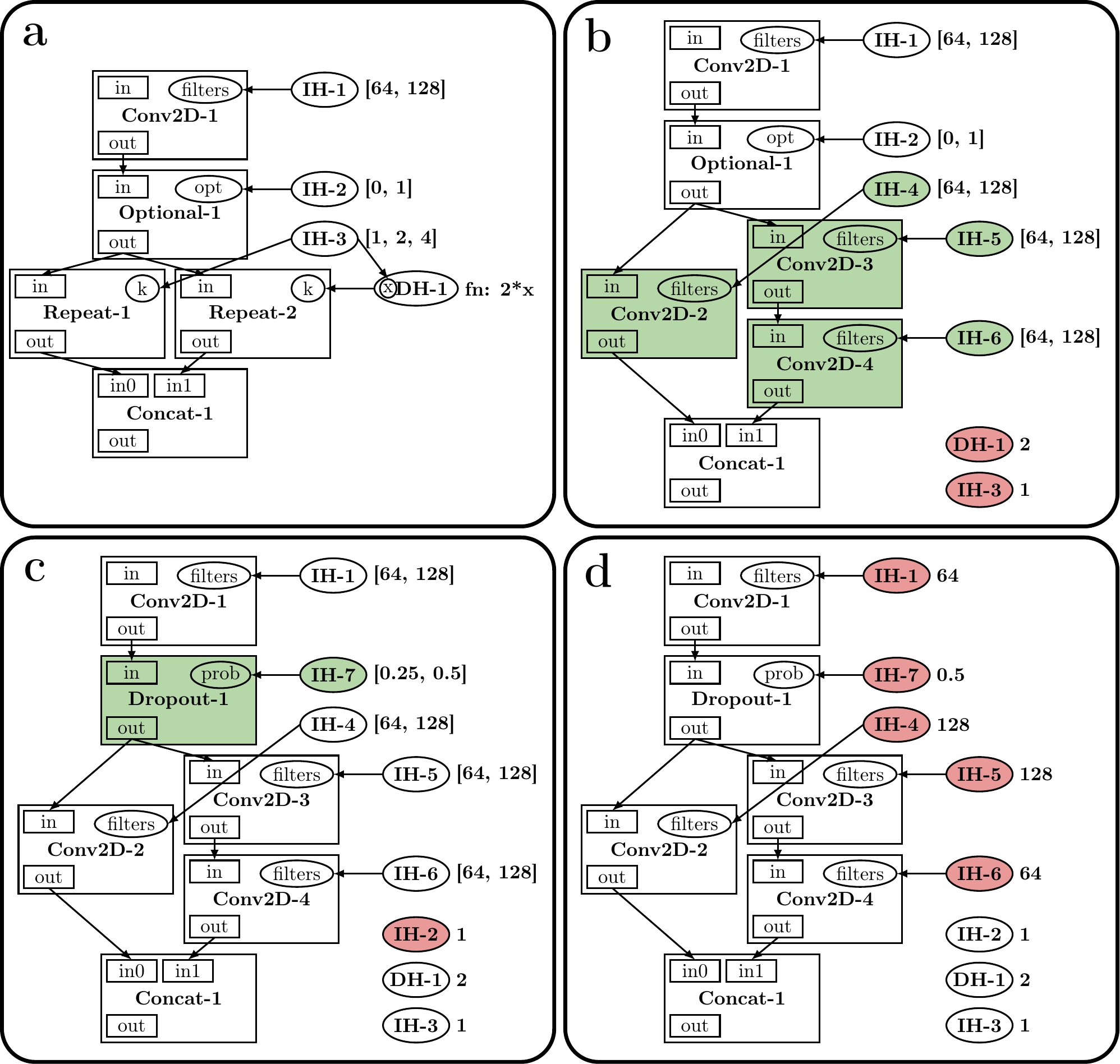}
\caption{Search space transitions for the search space in Figure~\ref{fig:example} (frame \code{a}) leading to a single architecture (frame \code{d}).
Modules and hyperparameters created since the previous frame are highlighted in green.
Hyperparameters assigned values since the previous frame are highlighted in red.
}
\label{fig:graph_transitions}
\end{figure}

We ground discussion textually, through code examples (Figure~\ref{fig:example}), and visually (Figure~\ref{fig:graph_transitions}) through an example search space.
There is a convolutional layer followed, optionally, by dropout with rate $0.25$ or
$0.5$.
After the optional dropout layer, there are two parallel chains of convolutional layers.
The first chain has length $1$, $2$, or $4$, and the second chain has double the length of the first.
Finally, the outputs of both chains are concatenated.
Each convolutional layer has $64$ or $128$ filters (chosen separately).
This search space has $25008$ distinct models.

Figure~\ref{fig:graph_transitions} shows a sequence of graph transitions for this search space.
\code{IH} and \code{DH} denote type identifiers for independent and dependent hyperparameters, respectively.
Modules and hyperparameters types are suffixed with a number to generate unique identifiers.
Modules are represented by rectangles that contain inputs, outputs, and properties.
Hyperparameters are represented by ellipses (outside of modules) and are associated to module properties (e.g., in frame \code{a}, \code{IH-1} is associated to \code{filters} of \code{Conv2D-1}).
To the right of an independent hyperparameter we show, before assignment, its set of possible values and, after assignment, its value (e.g., \code{IH-1} in frame \code{a} and in frame \code{d}, respectively).
Similarly, for a dependent hyperparameter we show, before assignment, the function that computes its value and, after assignment, its value (e.g., \code{DH-1} in frame \code{a} and in frame \code{b}, respectively).
Frame \code{a} shows the initial search space encoded in Figure~\ref{fig:example}.
From frame \code{a} to frame \code{b}, \code{IH-3} is assigned a value, triggering the value assignment for \code{DH-1} and the substitutions for \code{Repeat-1} and \code{Repeat-2}.
From frame \code{b} to frame \code{c}, \code{IH-2} is assigned value 1, creating \code{Dropout-1} and \code{IH-7} (its dropout rate hyperparameter).
Finally, from frame \code{c} to frame \code{d}, the five remaining independent hyperparameters are assigned values.
The search space in frame \code{d} has a single architecture that can be mapped to an implementation in a deep learning framework.

\section{Semantics and mechanics of the search space specification language}
\label{sec:implementation_of_language}

In this section, we formally describe the semantics and mechanics of our language and show how they can be used to implement search algorithms for arbitrary search spaces.

\subsection{Semantics}
\label{sec:notation}

\paragraph{Search space components} A search space $G$ has hyperparameters $H(G)$ and modules $M(G)$.
We distinguish between independent and dependent hyperparameters as $H_i(G)$ and $H_d(G)$, where $H(G) = H_i(G) \cup H_d(G)$ and $H_d(G) \cap H_i(G) = \emptyset$, and basic modules and substitution modules as $M_b(G)$ and $M_s(G)$, where $M(G) = M_b(G) \cup M_s(G)$ and $M_b(G) \cap M_s(G) = \emptyset$.

\paragraph{Hyperparameters}
We distinguish between hyperparameters that have been assigned a value and those that have not as $H_a(G)$ and $H_u(G)$.
We have $H(G) = H_u(G) \cup H_a(G)$ and $H_u(G) \cap H_a(G) = \emptyset$.
We denote the value assigned to an hyperparameter $h \in H_a(G)$ as $v_{(G), (h)} \in \mathcal X_{(h)}$, where $h \in H_a(G)$ and $\mathcal X_{(h)}$ is the set of possible values for $h$.
Independent and dependent hyperparameters are assigned values differently.
For $h \in H_i(G)$, its value is assigned directly from $\mathcal X_{(h)}$.
For $h \in H_d(G)$, its value is computed by evaluating a function $f_{(h)}$ for the values of $H(h)$, where $H(h)$ is the set of hyperparameters that $h$ depends on.
For example, in frame \code{a} of Figure~\ref{fig:graph_transitions}, for $h = \code{DH-1}$, $H(h) = \{ \code{IH-3} \}$.
In frame \code{b}, $H_a(G) = \{\code{IH-3}, \code{DH-1}\}$ and  $H_u(G) = \{\code{IH-1}, \code{IH-4}, \code{IH-5}, \code{IH-6}, \code{IH-2}\}$.

\paragraph{Modules}%
A module $m \in M(G)$ has inputs $I(m)$, outputs $O(m)$, and hyperparameters $H(m) \subseteq H(G)$ along with mappings assigning names local to the module to inputs, outputs, and hyperparameters, respectively, $\sigma_{(m), i} : S_{(m), i} \to I(m)$, $\sigma_{(m), o} : S_{(m), o} \to O(m)$, $\sigma_{(m), h} : S_{(m), h} \to H(m)$, where $S_{(m), i} \subset \Sigma^*$, $S_{(m), o} \subset \Sigma^*$, and $S_{(m), h} \subset \Sigma^*$, where $\Sigma^*$ is the set of all strings of alphabet $\Sigma$.
$S_{(m), i}$, $S_{(m), o}$, and $S_{(m), h}$ are, respectively, the local names for the inputs, outputs, and hyperparameters of $m$.
Both $\sigma_{(m), i}$ and $\sigma_{(m), o}$ are bijective, and therefore, the inverses $\sigma^{-1}_{(m), i} : I(m) \to S_{m,i}$ and $\sigma^{-1}_{(m),o} : O(m) \to S_{(m),o}$ exist and assign an input and output to its local name.
Each input and output belongs to a single module.
$\sigma_{(m), h}$ might not be injective, i.e., $|S_{(m), h}| \geq |H(m)|$.
A name $s \in S_{(m), h}$ captures the local semantics of $\sigma_{(m), h}(s)$ in $m \in M(G)$ (e.g., for a convolutional basic module, the number of filters or the kernel size).
Given an input $i \in I(M(G))$, $m(i)$ recovers the module that $i$ belongs to (analogously for outputs).
For $m \neq m'$, we have $I(m) \cap I(m') = \emptyset$ and $O(m) \cap O(m') = \emptyset$,
but there might exist $m, m' \in M(G)$ for which $H(m) \cap H(m') \neq \emptyset$, i.e.,
two different modules might share hyperparameters but inputs and outputs belong to a single module.
We use shorthands $I(G)$ for $I(M(G))$ and $O(G)$ for $O(M(G))$.
For example, in frame \code{a} of Figure~\ref{fig:graph_transitions}, for $m = \code{Conv2D-1}$ we have: $I(m) = \{\code{Conv2D-1.in}\}$, $O(m) = \{\code{Conv2D-1.out}\}$, and $H(m) = \{ \code{IH-1} \}$; $S_{(m), i} = \{\code{in}\}$ and $\sigma_{(m), i}(\code{in}) = \code{Conv2D-1.in}$ ($\sigma_{(m), o}$ and $\sigma_{(m), h}$ are similar); $m(\code{Conv2D-1.in}) = \code{Conv2D-1}$.
Output and inputs are identified by the global name of their module and their local name within their module joined by a dot, e.g.. \code{Conv2D-1.in}

\paragraph{Connections between modules}
Connections between modules in $G$ are represented through the set of directed edges $E(G) \subseteq O(G) \times I(G)$ between outputs and inputs of modules in $M(G)$.
We denote the subset of edges involving inputs of a module $m \in M(G)$ as $E_i(m)$, i.e., $E_i(m) = \{(o, i) \in E(G) \mid i \in  I(m)\}$.
Similarly, for outputs, $E_o(m) = \{(o, i) \in E(G) \mid o \in O(m)\}$.
We denote the set of edges involving inputs or outputs of $m$ as $E(m) = E_i(m) \cup E_o(m)$.
In frame $a$ of Figure~\ref{fig:graph_transitions},
For example, in frame \code{a} of Figure~\ref{fig:graph_transitions}, $E_i(\code{Optional-1}) = \{ (\code{Conv2D-1.out}, \code{Optional-1.in} ) \}$ and  $E_o(\code{Optional-1}) = \{ (\code{Optional-1.out}, \code{Repeat-1.in}), (\code{Optional-1.out}, \code{Repeat-2.in}) \}$.

\paragraph{Search spaces}
We denote the set of all possible search spaces as $\mathcal G$.
For a search space $G \in \mathcal G$, we define $\mathcal R(G) = \{G' \in \mathcal G \mid G _1, \ldots, G_m \in \mathcal G ^m, G_{k + 1} = \code{Transition}( G_k, h, v ), h \in H_i(G_k) \cap H_u(G_k), v \in \mathcal X_{(h)}, \forall k \in [m], G_1 = G, G_m = G'\}$, i.e., the set of reachable search spaces through a sequence of value assignments to independent hyperparameters (see Algorithm~\ref{alg:graph_transition} for the description of \code{Transition}).
We denote the set of terminal search spaces as $\mathcal T \subset \mathcal G$, i.e.  $\mathcal T = \{G \in \mathcal G \mid H_i(G) \cap H_u(G) = \emptyset\}$.
We denote the set of terminal search spaces that are reachable from $G \in \mathcal G$ as $\mathcal T(G) = \mathcal R(G) \cap \mathcal T$.
In Figure~\ref{fig:graph_transitions}, if we let $G$ and $G'$ be the search spaces in frame \code{a} and \code{d}, respectively, we have $G' \in \mathcal T(G)$.

\subsection{Mechanics}
\label{ssec:mechanics}

\begin{figure}[tbp]
        \centering
\begin{minipage}{0.63\textwidth}
    \begin{algorithm}[H]
    \small
        \SetKwRepeat{Do}{do}{while}%
        \KwIn{$G, h \in H_i(G) \cap H_u(G), v \in \mathcal X_{(h)}$}
        $v_{(G), (h)} \leftarrow v$ \\
        \Do{$\tilde H_d(G) \neq \emptyset$ or $\tilde M_s(G) \neq \emptyset$}{
             $\tilde H_d(G) = \{ h \in H_d(G) \cap H_u(G) \mid H_u(h) = \emptyset \}$ \\
            \For{$h \in \tilde H_d(G)$}{
                $n \leftarrow |S_{(h)}|$ \\
                Let $S_{(h)} = \{s_1, \ldots, s_n\}$ with $s_1 < \ldots < s_n$ \\
                $v_{(G), (h)} \leftarrow f_{(h)}(v_{G, \sigma_{(h)}(s_1)}, \ldots, v_{G, \sigma_{(h)}(s_n)})$
            }
             $\tilde M_s(G) = \{ m \in M_s(G) \mid H_u(m) = \emptyset \}$ \\
            \For{$m \in \tilde M_s(G)$}{
                $n \leftarrow |S_{(m), h}|$ \\
                Let $S_{(m), h} = \{s_1, \ldots, s_n\}$ with $s_1 < \ldots < s_n$ \\
                $(G_m, \sigma_i, \sigma_o) = f_{(m)}(v_{G, \sigma_{(m), h}(s_1)}, \ldots, v_{G, \sigma_{(m), h}(s_n)})$\\
                $E_{i} = \{ (o, i') \mid
                    (o, i) \in E_i(m),
                    i' = \sigma_{i}( \sigma^{-1}_{(m), i}(i) )  \} $ \\
                $E_o = \{ (o', i) \mid
                    (o, i) \in E_o(m),
                    o' = \sigma_{o}( \sigma^{-1}_{(m), o}(o) ) \} $ \\
                $E(G) \leftarrow \left( E(G) \setminus E(m) \right) \cup \left( E_i \cup E_o \right)$ \\
                $M(G) \leftarrow \left( M(G) \setminus \{ m \} \right) \cup M(G_m)$ \\
                $H(G) \leftarrow H(G) \cup H(G_m)$
            }
            }
        \Return{$G$}
    \caption{\code{Transition}}
    \label{alg:graph_transition}
    \end{algorithm}
\end{minipage}
 \hfill
\begin{minipage}{0.35\textwidth}
    \centering
    \begin{algorithm}[H]
    \small
        \KwIn{$G, \sigma_o : S_o \to O_u(G)$}
        $M_q \leftarrow \code{OrderedModules}(G, \sigma_o) $ \\
        $H_q \leftarrow [\,]$ \\
        \For{$m \in M_q$}{
            $n = |S_{(m), h}|$ \\
            Let $S_{(m), h} = \{s_1, \ldots, s_{ n } \}$ with $s_1 < \ldots < s_{n}$.\\
            \For{$j \in [n]$}{
                $h \leftarrow \sigma_{(m), h}(s_j)$ \\
                \If{$h \notin H_q$}{
                    $H_q \leftarrow H_q + [h]$\\
                }
            }
        }
        \For{$h \in H_q$}{
            \If{$h \in H_d(G)$}{
                $n \leftarrow |S_{(h)}|$ \\
                Let $S_{(h)} = \{s_1, \ldots, s_n\}$ with $s_1 < \ldots < s_n$ \\
                \For{$j \in [n]$}{
                    $h' \leftarrow \sigma_{(h)}(s_j)$ \\
                    \If{$h' \notin H_q$}{
                        $H_q \leftarrow H_q + [h']$\\
                    }
                }
            }
        }
        \Return{$H_q$}
    \caption{\code{OrderedHyperps}}
    \label{alg:hyperp_traversal}
    \end{algorithm}
\end{minipage}
        \label{fig:traversals}
    \caption{
    \textit{Left:} \code{Transition} assigns a value to an independent hyperparameter and resolves assignments to dependent hyperparameters (line 3 to 7) and substitutions (line 8 to 17) until none are left (line 18).
    \textit{Right:} \code{OrderedHyperps} returns $H(G)$ sorted according to a unique order.
    Adds the hyperparameters that are immediately reachable from modules (line 1 to 9), and then traverses the dependencies of the dependent hyperparameters to find additional hyperparameters (line 10 to 17).
    }
    \label{fig:main_algorithms}
\end{figure}

\paragraph{Search space transitions}

A search space $G \in \mathcal G$ encodes a set of architectures (i.e., those in $\mathcal T(G)$).
Different architectures are obtained through different sequences of value assignments leading to search spaces in $\mathcal T(G)$.
Graph transitions result from value assignments to independent hyperparameters.
Algorithm~\ref{alg:graph_transition} shows how the search space $G' = \code{Transition}(G, h, v)$ is computed, where $h \in H_i(G) \cap H_u(G)$ and $v \in \mathcal X_{(h)}$.
Each transition leads to progressively smaller search spaces (i.e., for all $G \in \mathcal G, G' = \code{Transition}(G, h, v)$ for $h \in H_i(G) \cap H_u(G)$ and $v \in \mathcal X_{(h)}$, then $\mathcal R(G') \subseteq \mathcal R(G)$).
A search space $G' \in \mathcal T(G)$ is reached once there are no independent hyperparameters left to assign values to, i.e., $H_i(G) \cap H_u(G) = \emptyset$.
For $G' \in \mathcal T(G)$, $M_s(G') = \emptyset$, i.e., there are only basic modules left.
For search spaces $G \in \mathcal G$ for which $M_s(G) = \emptyset$, we have $M(G') = M(G)$ (i.e., $M_b(G') = M_b(G)$) and $H(G') = H(G)$ for all $G' \in \mathcal R(G)$, i.e., no new modules and hyperparameters are created as a result of graph transitions.
Algorithm~\ref{alg:graph_transition} can be implemented efficiently by checking whether assigning a value to $h \in H_i(G) \cap H_u(G)$ triggered substitutions of neighboring modules or value assignments to neighboring hyperparameters.
For example, for the search space $G$ of frame \code{d} of Figure~\ref{fig:graph_transitions}, $M_s(G) = \emptyset$.
Search spaces $G$, $G'$, and $G''$ for frames \code{a}, \code{b}, and \code{c}, respectively, are related as $G' = \code{Transition}(G, \code{IH-3}, 1)$ and $G'' = \code{Transition}(G', \code{IH-2}, 1)$.
For the substitution resolved from frame \code{b} to frame \code{c}, for $m = \code{Optional-1}$, we have $\sigma_i(\code{in}) = \code{Dropout-1.in}$ and $\sigma_o(\code{out}) = \code{Dropout-1.out}$ (see line 12 in Algorithm~\ref{alg:graph_transition}).

\paragraph{Traversals over modules and hyperparameters}

Search space traversal is fundamental to provide the interface to search spaces that search algorithms rely on (e.g., see Algorithm~\ref{alg:random_search}) and to automatically map terminal search spaces to their runnable computational graphs (see Algorithm~\ref{alg:graph_propagation} in Appendix~\ref{app:mechanics_details}).
For $G \in \mathcal G$, this iterator is implemented by using Algorithm~\ref{alg:hyperp_traversal} and keeping only the hyperparameters in $H_u(G) \cap H_i(G)$.
The role of the search algorithm (e.g., see Algorithm~\ref{alg:random_search}) is to recursively assign values to hyperparameters in $H_u(G) \cap H_i(G)$ until a search space $G' \in \mathcal T(G)$ is reached.
Uniquely ordered traversal of $H(G)$ relies on uniquely ordered traversal of $M(G)$.
(We defer discussion of the module traversal to Appendix~\ref{app:mechanics_details}, see Algorithm~\ref{alg:module_traversal}.)

\paragraph{Architecture instantiation}

A search space $G \in \mathcal T$ can be mapped to a domain implementation (e.g. computational graph in Tensorflow~\cite{abadi2016tensorflow} or PyTorch~\cite{paszke2017automatic}).
Only fully-specified basic modules are left in a terminal search space $G$ (i.e., $H_u(G) = \emptyset$ and $M_s(G) = \emptyset$).
The mapping from a terminal search space to its implementation relies on graph traversal of the modules according to the topological ordering of their dependencies (i.e., if $m'$ connects to an output of $m$, then $m'$ should be visited after $m$).
\begin{wrapfigure}[19]{r}{0.5\textwidth}
\hspace{0.2cm}
  \begin{algorithm}[H]
  \small
      \KwIn{$G, \sigma_o : S_o \to O_u(G), k$}
      $r_{\text{best}} \leftarrow - \infty$ \\
      \For{$j \in [k]$}{
          $G' \leftarrow G$  \\
          \While{$G' \notin \mathcal T$}{
              $H_q \leftarrow \code{OrderedHyperps}(G', \sigma_o)$ \\
              \For{$h \in H_q$}{
                  \If{$h \in H_u(G') \cap H_i(G')$}{
                      $v \sim \code{Uniform}(\mathcal X_{(h)})$ \\
                      $G' \leftarrow \code{Transition}(G', h, v)$
                  }
              }
          }
          $r \leftarrow \code{Evaluate}(G')$ \\
          \If{$r > r_{\text{best}}$}{
              $r_{\text{best}} \leftarrow r$ \\
              $G_{\text{best}} \leftarrow G'$
          }
        }
        \Return{$G_{\text{best}}$}
  \caption{Random search.}
  \label{alg:random_search}
  \end{algorithm}
  \caption{Assigns a value uniformly at random (line 8) for each independent hyperparameter (line 7) in the search space until a terminal search space is reached (line 4).}
  \end{wrapfigure}
Appendix~\ref{app:mechanics_details} details this graph propagation process (see Algorithm~\ref{alg:graph_propagation}).
For example, it is simple to see how the search space of frame \code{d} of Figure~\ref{fig:graph_transitions} can be mapped to an implementation.

\subsection{Supporting search algorithms}
\label{ssec:supporting_search_algorithms}

Search algorithms interface with search spaces through ordered iteration over unassigned independent hyperparameters (implemented with the help of Algorithm~\ref{alg:hyperp_traversal}) and value assignments to these hyperparameters (which are resolved with Algorithm~\ref{alg:graph_transition}).
Algorithms are run for a fixed number of evaluations $k \in \mathbb N$, and return the best architecture found.
The iteration functionality in Algorithm~\ref{alg:hyperp_traversal} is independent of the search space and therefore can be used to expose search spaces to search algorithms.
We use this decoupling to mix and match search spaces and search algorithms without implementing each pair from scratch (see Section~\ref{sec:experiments}).

\section{Experiments}
\label{sec:experiments}

We showcase the modularity and programmability of our language by running experiments that rely on decoupled of search spaces and search algorithms.
The interface to search spaces provided by the language makes it possible to reuse implementations of search spaces and search algorithms.

\subsection{Search space experiments}
\label{ssec:search_space_experiments}

\begin{wraptable}{r}{0.4\textwidth}
    \centering
        \caption{Test results for search space experiments.}
    \begin{tabular}{lc}
        \toprule
        Search Space  & Test Accuracy \\
        \midrule
        Genetic~\cite{xie2017genetic}       & 90.07  \\
        Flat~\cite{liu2017hierarchical}     & 93.58 \\
        Nasbench~\cite{ying2019nasbench}    & 94.59 \\
        Nasnet~\cite{zoph2018learning}      & 93.77 \\
        \bottomrule
    \end{tabular}
    \label{tab:ss_test_results}
    \end{wraptable}

\begin{figure}[tbp]
    \centering
    \begin{subfigure}{.48\textwidth}
        \centering
        \includegraphics[width=\textwidth]{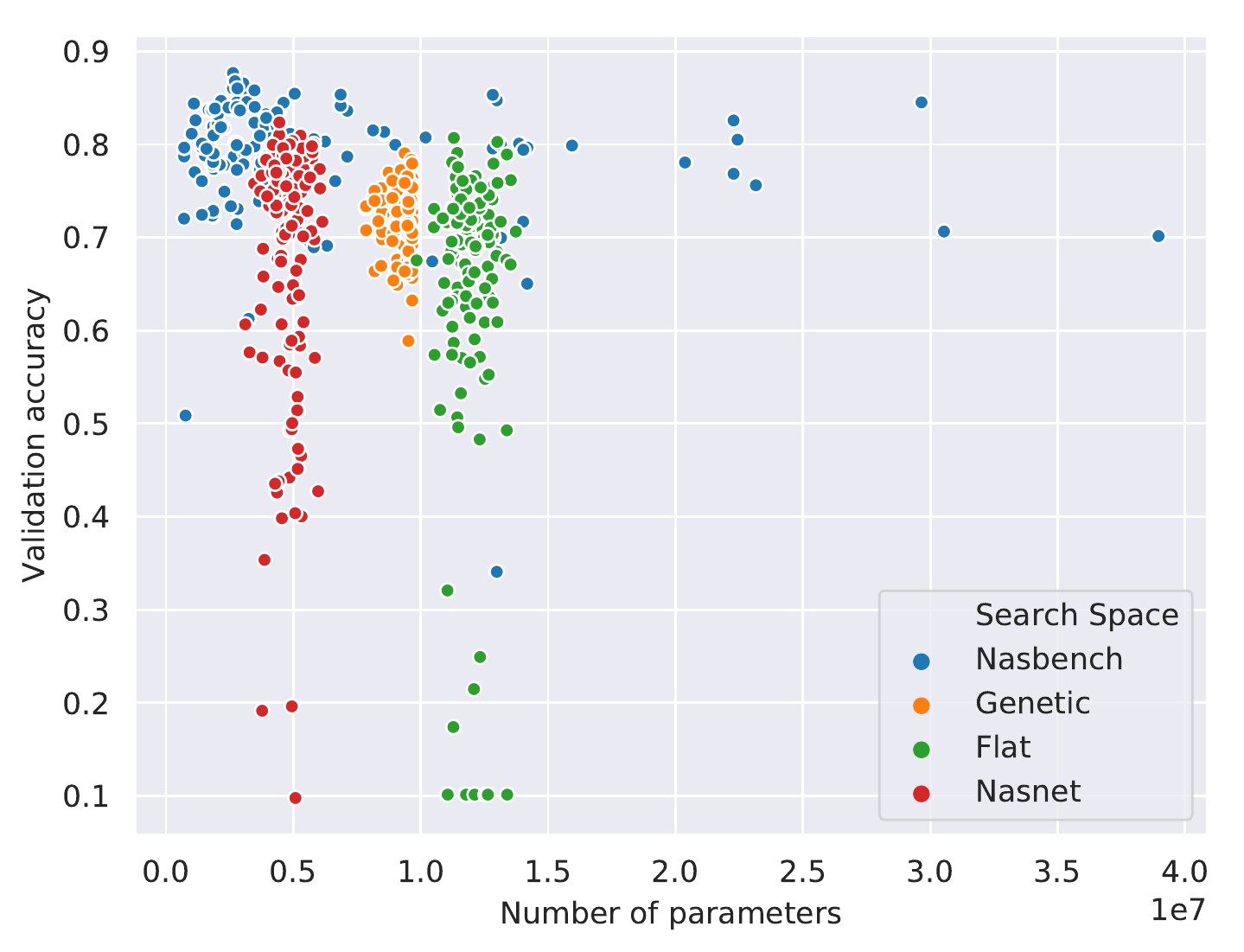}
        \label{fig:ss_num_params_vs_val_acc}
    \end{subfigure}
    \hspace{.04in}
    \begin{subfigure}{.48\textwidth}
        \centering
        \includegraphics[width=\textwidth]{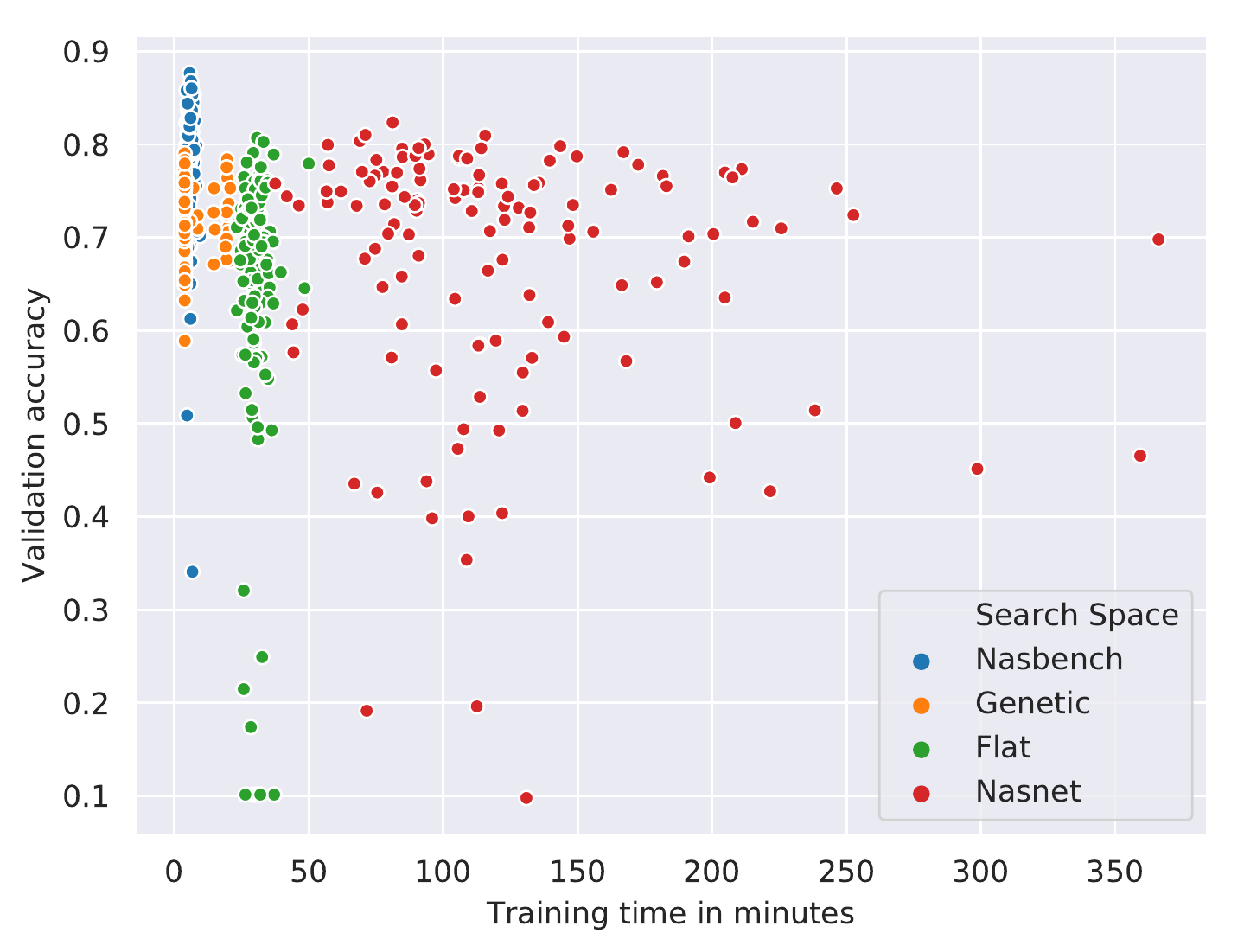}
        \label{fig:ss_time_vs_val_acc}
    \end{subfigure}
    \caption{Results for the architectures sampled in the search space experiments.
    \emph{Left}: Relation between number of parameters and validation accuracy at 25 epochs.
    \emph{Right}: Relation between time to complete 25 epochs of training and validation accuracy.}
    \label{fig:search_space_experiments_plots}
\end{figure}

We vary the search space and fix the search algorithm and the evaluation method.
We refer to the search spaces we consider as Nasbench~\cite{ying2019nasbench}, Nasnet~\cite{zoph2018learning}, Flat~\cite{liu2017hierarchical}, and Genetic~\cite{xie2017genetic}.
For the search phase, we randomly sample $128$ architectures from each search space and train them for $25$ epochs with Adam with a learning rate of $0.001$.
The test results for the fully trained architecture with the best validation accuracy are reported in Table~\ref{tab:ss_test_results}.
These experiments provide a simple characterization of the search spaces in terms of the number of parameters, training times, and validation performances at $25$ epochs of the architectures in each search space (see Figure~\ref{fig:search_space_experiments_plots}).
Our language makes these characterizations easy due to better modularity (the implementations of the search space and search algorithm are decoupled) and programmability (new search spaces can be encoded and new search algorithms can be developed).

\subsection{Search algorithm experiments}
\label{ssec:search_algorithm_experiments}

    \begin{wraptable}{r}{0.4\textwidth}
        \centering
          \caption{Test results for search algorithm experiments.} 
        \begin{tabular}{lc}
          \toprule
          Search algorithm     & Test Accuracy \\
          \midrule
          Random                                  & $91.61\pm0.67$ \\
          MCTS~\cite{browne2012survey}           & $91.45 \pm 0.11$ \\
          SMBO~\cite{negrinho2017deeparchitect}  & $91.93 \pm 1.03$ \\
          Evolution~\cite{real2018regularized}   & $91.32	\pm 0.50$ \\
          \bottomrule
        \end{tabular}
        \label{tab:se_test_results}
      \end{wraptable}

\begin{figure}[tbp]
    \centering
    \begin{subfigure}{.48\textwidth}
        \centering
        \includegraphics[width=\textwidth]{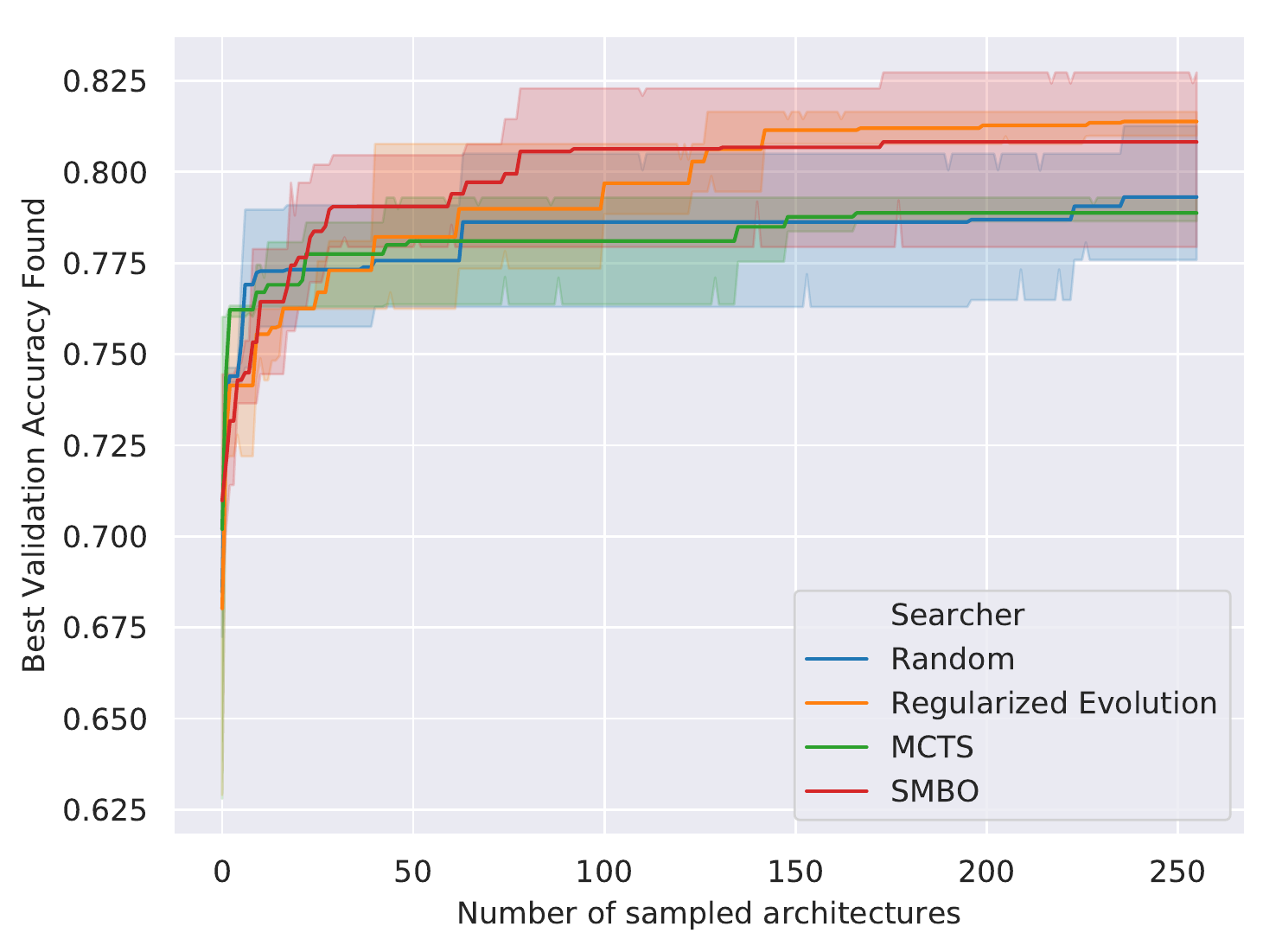}
        \label{fig:se:best_val_vs_id}
    \end{subfigure}
    \hspace{.04in}
    \begin{subfigure}{.48\textwidth}
        \centering
        \includegraphics[width=\textwidth]{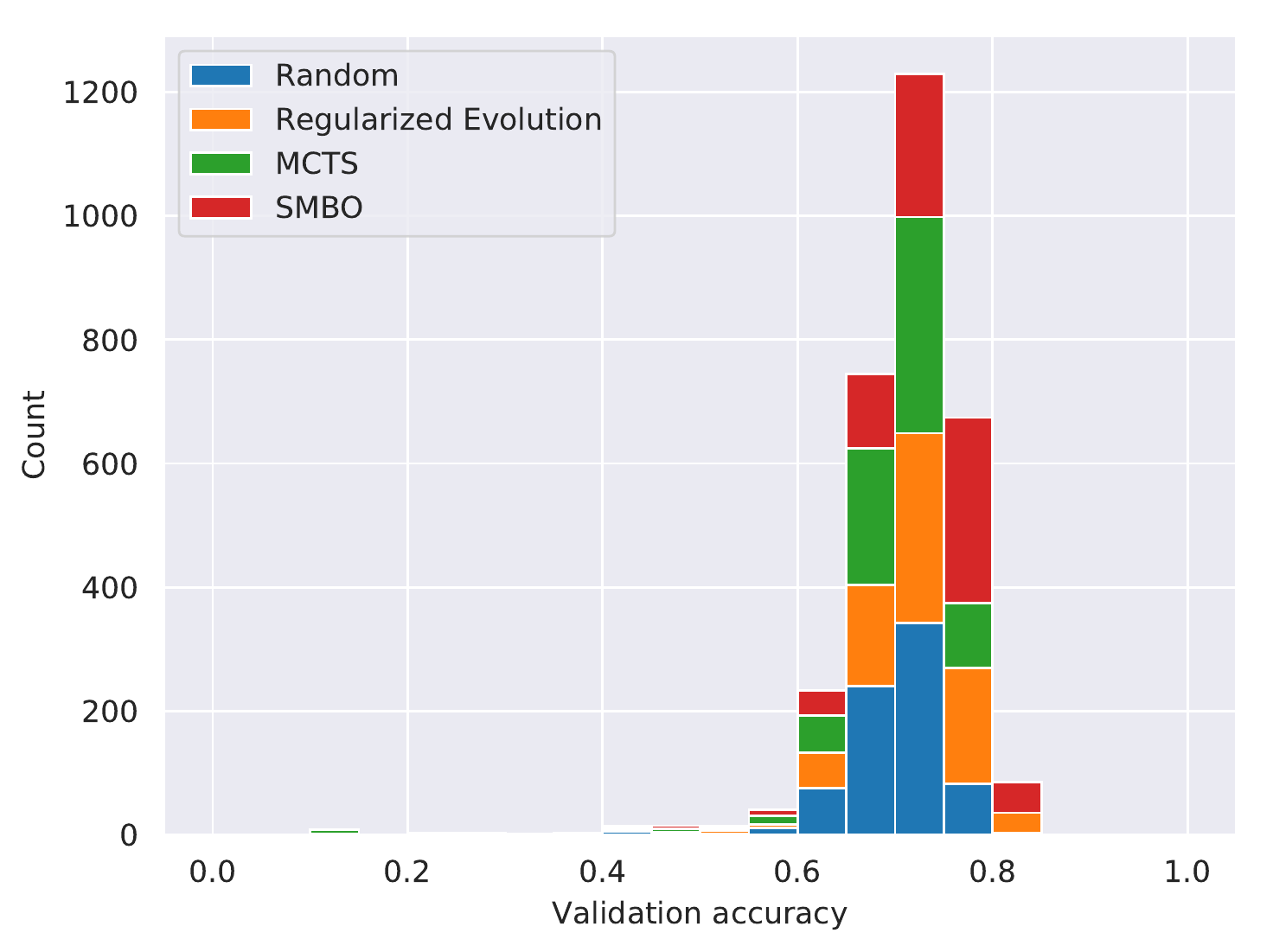}
        \label{fig:se:val_acc_hist}
    \end{subfigure}
    \caption{
        Results for search algorithm experiments.
        \emph{Left}: Relation between the performance of the best architecture found and the number of architectures sampled.
        \emph{Right}: Histogram of validation accuracies for the architectures encountered by each search algorithm.}
    \label{fig:search_algorithm_plots}
\end{figure}

We evaluate search algorithms by running them on the same search space.
We use the Genetic search space~\cite{xie2017genetic} for these experiments as Figure~\ref{fig:search_space_experiments_plots} shows its architectures train quickly and have substantially different validation accuracies.
We examined the performance of four search algorithms: random, regularized evolution, sequential model based optimization (SMBO), and Monte Carlo Tree Search (MCTS).
Random search uniformly samples values for independent hyperparameters (see Algorithm~\ref{alg:random_search}).
Regularized evolution~\cite{real2018regularized} is an evolutionary algorithm that mutates the best performing member of the population and discards the oldest.
We use population size $100$ and sample size $25$.
For SMBO~\cite{negrinho2017deeparchitect}, we use a linear surrogate function to predict the validation accuracy of an architecture from its features (hashed modules sequences and hyperparameter values).
For each architecture requested from this search algorithm, with probability $0.1$ a randomly specified architecture is returned; otherwise it evaluates $512$ random architectures with the surrogate model and returns the one with the best predicted validation accuracy.
MCTS~\cite{browne2012survey,negrinho2017deeparchitect} uses the Upper Confidence Bound for Trees (UCT) algorithm with the exploration term of $0.33$.
Each run of the search algorithm samples $256$ architectures that are trained for $25$ epochs with Adam with a learning rate of $0.001$.
We ran three trials for each search algorithm.
See Figure~\ref{fig:search_algorithm_plots} and Table~\ref{tab:se_test_results} for the results.
By comparing Table~\ref{tab:ss_test_results} and Table~\ref{tab:se_test_results}, we see that the choice of search space had a much larger impact on the test accuracies observed than the choice of search algorithm.
See Appendix~\ref{app:additional_results} for more details.

\section{Conclusions}
\label{sec:conclusion}

We design a language to encode search spaces over architectures to improve the programmability and modularity of architecture search research and practice.
Our language allows us to decouple the implementations of search spaces and search algorithms.
This decoupling enables to mix-and-match search spaces and search algorithms without having to write each pair from scratch.
We reimplement search spaces and search algorithms from the literature and compare them under the same conditions.
We hope that decomposing architecture search experiments through the lens of our language will lead to more reusable and comparable architecture search research.

\section{Acknowledgements}
\label{sec:acks}

We thank the anonymous reviewers for helpful comments and suggestions.
We thank Graham Neubig, Barnabas Poczos, Ruslan Salakhutdinov, Eric Xing, Xue Liu, Carolyn Rose, Zhiting Hu, Willie Neiswanger, Christoph Dann, Kirielle Singajarah, and Zejie Ai for helpful discussions.
We thank Google for generous TPU and GCP grants.
This work was funded in part by NSF grant IIS 1822831.

{
\small
    \bibliography{deep_architect}
    \bibliographystyle{unsrt}
}

\newpage
\appendix
\clearpage

\section{Additional details about language components}
\label{app:components}

\paragraph{Independent hyperparameters}
\label{app:independent_hyperparameters}

\begin{wrapfigure}{r}{0.5\textwidth}
\begin{lstlisting}[style = Python]
h_filters = D([32, 64, 128])
h_stride = D([1])
conv_fn = lambda h_kernel_size: conv2d(
    h_filters, h_stride, h_kernel_size)
(c1_inputs, c1_outputs) = conv_fn(D([1, 3, 5]))
(c2_inputs, c2_outputs) = conv_fn(D([1, 3, 5]))
c1_outputs["out"].connect(c2_inputs["in"])
\end{lstlisting}
\caption{Search space with two convolutions in series.
The number of filters is the same for both, while the kernel sizes are chosen separately.}
\label{fig_app:example1}
\end{wrapfigure}

An hyperparameter can be shared by instantiating it and using it in multiple modules.
For example, in Figure~\ref{fig_app:example1}, \code{conv\_fn} has access to \code{h\_filters} and \code{h\_stride} through a closure and uses them in boths calls.
There are $27$ architectures in this search space (corresponding to the possible choices for the number of filters, stride, and kernel size).
The output of the first convolution is connected to the input of the second through the call to \code{connect} (line 7).

\paragraph{Dependent hyperparameters}
\label{app:Dependent_hyperparameters}

\begin{wrapfigure}{r}{0.5\textwidth}
\begin{lstlisting}[style = Python]
h_filters_lst = [D([32, 64, 128])]
h_factor = D([1, 2, 4])
h_stride = D([1])
io_lst = []
for i in range(3):
    h = h_filters_lst[i]
    (inputs, outputs) = conv2d(h, h_stride,
                                D([1, 3, 5]))
    io_lst.append((inputs, outputs))
    if i > 0:
        io_lst[i - 1][1]["out"].connect(
            io_lst[i][0]["in"])
    if i < 2:
        h_next = DependentHyperparameter(
            lambda x, y: x * y,
            {"x": h, "y": h_factor})
        h_filters_lst.append(h_next)
\end{lstlisting}
    \caption{Search space with three convolutions in series.
    The number of filters of an inner convolution is a multiple of the number of filters of the previous convolution.
    The multiple is chosen through an hyperparameter (\code{h\_factor}).}
    \label{fig_app:example2}
    \end{wrapfigure}

Chains (or general directed acyclic graphs) involving dependent and independent hyperparameters are valid.
The search space in Figure~\ref{fig_app:example2} has three convolutional modules in series.
Each convolutional module shares the hyperparameter for the stride, does not share the hyperparameter for the kernel size, and relates the hyperparameters for the number of filters via a chain of dependent hyperparameters.
Each dependent hyperparameter depends on the previous hyperparameter and on the multiplier hyperparameter.
This search space has $243$ distinct architectures

Encoding this search space in our language might not seem advantageous when compared to encoding it in an hyperparameter optimization tool.
Similarly to ours, the latter requires defining hyperparameters for the multiplier, the initial number of filters, and the three kernel sizes (chosen separately).
Unfortunately, the encoding by itself tells us nothing about the mapping from hyperparameter values to implementations---the expert must write separate code for this mapping and change it when the search space changes.
By contrast, in our language the expert only needs to write the encoding for the search space---the mapping to implementations is induced automatically from the encoding.

\begin{figure}[htbp]
    \centering
    \begin{minipage}{\textwidth}
    \begin{minipage}{0.47\textwidth}
    \begin{lstlisting}[style = Python]
def dense(h_units):
    def compile_fn(di, dh):
        m =  tf.layers.Dense(dh['units'])
        def forward_fn(di):
            return {"out": m(di["in"])}
        return forward_fn
    name_to_hyperp = {'units': h_units}
    return siso_tensorflow_module(
        'Affine', compile_fn, name_to_hyperp, scope)
    \end{lstlisting}
    \end{minipage}
    \hspace{0.55cm}
\begin{minipage}{0.47\textwidth}
\begin{lstlisting}[style = Python]
def conv2d(h_num_filters, h_filter_width, h_stride):
    def compile_fn(di, dh):
        conv_op = tf.layers.Conv2D(
            dh['num_filters'],
            (dh['filter_width'],) * 2,
            (dh['stride'],) * 2,
            padding='SAME')
        def forward_fn(di):
            return {'out': conv_op(di['in'])}
        return forward_fn
    return siso_tensorflow_module(
        'Conv2D', compile_fn, {
            'num_filters': h_num_filters,
            'filter_width': h_filter_width,
            'stride': h_stride
        })
\end{lstlisting}
\end{minipage}
\end{minipage}
\caption{
    Examples of basic modules in our implementation resulting from wrapping Tensorflow operations.
    \emph{Left:} Affine basic module with an hyperparameter for the number of units.
    \emph{Right:} Convolutional basic module with hyperparameters for the number of filters, filter size, and stride. }
\label{fig_app:basic_modules}
\end{figure}

\paragraph{Basic Modules}
\label{app:basic_modules}
Deep learning layers can be easily wrapped as basic modules.
For example, a dense layer can be wrapped as a single-input single-output module with one hyperparameter for the number of units (see left of Figure~\ref{fig_app:basic_modules}).
A convolutional layer is another example of a single-input single-output module (see right of Figure~\ref{fig_app:basic_modules}).
The implementation of \code{conv2d} relies on \code{siso\_tensorflow\_module} for wrapping Tensorflow-specific aspects (see Appendix~\ref{sec:domain_support} for a discussion on how to support different domains).
\code{conv2d} depends on hyperparameters for \code{num\_filters}, \code{filter\_width}, and \code{stride}.
The key observation is that a basic module generates its implementation (calls to \code{compile\_fn} and then \code{forward\_fn}) only after its hyperparameter values have been assigned and it has values for its inputs.
The values of the inputs and the hyperparameters are available in the dictionaries \code{di} and \code{dh}, respectively.
\code{conv2d} returns a module as \code{(inputs, outputs)} (these are analogous to $\sigma_i$ and $\sigma_h$ on line of 12 of Algorithm~\ref{alg:graph_transition}).
Instantiating the computational graph relies on \code{compile\_fn} and \code{forward\_fn}.
\code{compile\_fn} is called a single time, e.g., to instantiate the parameters of the basic module.
\code{forward\_fn} can be called multiple times to create the computational graph (in static frameworks such as Tensorflow) or to evaluate the computational graph for specific data (e.g., in dynamic frameworks such as PyTorch).
Parameters instantiated in \code{compile\_fn} are available to \code{forward\_fn} through a closure.

\begin{figure}[htbp]
    \centering
    \begin{minipage}{\textwidth}
    \begin{minipage}{0.47\textwidth}
\begin{lstlisting}[style = Python]
def mimo_or(fn_lst, h_or, input_names,
        output_names, scope=None, name=None):
    def substitution_fn(dh):
        return fn_lst[dh["idx"]]()

    return substitution_module(
        _get_name(name, "Or"),
        substitution_fn,
        {'idx': h_or},
        input_names, output_names, scope)
    \end{lstlisting}
\begin{lstlisting}[style = Python]
def siso_repeat(fn, h_num_repeats,
        scope=None, name=None):
    def substitution_fn(dh):
        assert dh["num_reps"] > 0
        return siso_sequential([fn()
            for _ in range(dh["num_reps"])])

    return substitution_module(
        _get_name(name, "SISORepeat"),
        substitution_fn,
        {'num_reps': h_num_repeats},
        ['in'], ['out'], scope)
\end{lstlisting}
    \end{minipage}
    \hspace{0.55cm}
\begin{minipage}{0.47\textwidth}
        \begin{lstlisting}[style = Python]
def siso_split_combine(fn, combine_fn,
        h_num_splits, scope=None, name=None):
    def substitution_fn(dh):
        inputs_lst, outputs_lst = zip(*[fn()
            for _ in range(dh["num_splits"])])
        c_inputs, c_outputs = combine_fn(
            dh["num_splits"])

        i_inputs, i_outputs = identity()
        for i in range(dh["num_splits"]):
            i_outputs['out'].connect(
                inputs_lst[i]['in'])
            c_inputs['in' + str(i)].connect(
                outputs_lst[i]['out'])
        return i_inputs, c_outputs

    return substitution_module(
        _get_name(name, "SISOSplitCombine"),
        substitution_fn,
        {'num_splits': h_num_splits},
        ['in'], ['out'], scope)
\end{lstlisting}
    \end{minipage}
    \end{minipage}
    \caption{Example substitution modules implemented in our framework.
    \emph{Top left:} \code{mimo\_or} chooses between a list of functions returning search spaces.
    \emph{Bottom left:} Creates a series connection of the search space returned by \code{fn} some number of times (determined by \code{h\_num\_repeats}).
    \emph{Right:} Creates a search space with a number (determined by \code{h\_num\_splits}) of single-input single-output parallel search spaces created by \code{fn} that are then combined into the search space created by \code{combine\_fn}.
    }
    \label{fig_app:substitution_modules}
    \end{figure}

\paragraph{Substitution modules}
\label{app:substitution_modules}
Substitution modules encode local structural transformations of the search space that are resolved once all their hyperparameters have been assigned values (see line 12 in Algorithm~\ref{alg:graph_transition}).
Consider the implementation of \code{mimo\_or} (i.e., mimo stands for multi-input, multi-output) in Figure~\ref{fig_app:substitution_modules} (top left).
We make substantial use of higher-order functions and closures in our language implementation.
For example, to implement a specific \code{or} substitution module, we only need to provide a list of functions that return search spaces.
Arguments that the functions would need to carry are accessed through the closure or through argument binding\footnote{This is often called a thunk in programming languages.}.
\code{mimo\_or} has an hyperparameter for which subsearch space function to pick (\code{h\_idx}).
Once \code{h\_idx} is assigned a value, \code{substitution\_fn} is called, returning a search space as \code{(inputs, outputs)} where \code{inputs} and \code{outputs} are $\sigma_i$ and $\sigma_o$ mentioned on line 12 of Algorithm~\ref{alg:graph_transition}.
Using mappings of inputs and outputs is convenient because it allow us to treat modules and search spaces the same (e.g., when connecting search spaces).
The other substitution modules in Figure~\ref{fig_app:substitution_modules} use \code{substitution\_fn} similarly.

\begin{wrapfigure}[6]{r}{0.5\textwidth}
    \begin{lstlisting}[style = Python]
def substitution_module(name, name_to_hyperp,
        substitution_fn, input_names, output_names):
    \end{lstlisting}
    \caption{Signature of the helper used to create substitution modules.
    }
    \label{fig_app:substitution_module}
    \end{wrapfigure}

Figure~\ref{fig_app:substitution_module} shows the signature of the wrapper function to easily create substitution modules.
All information about what subsearch space should be generated upon substitution is delegated to \code{substitution\_fn}.
Compare this to signature of \code{keras\_module} for Keras basic modules in Figure~\ref{fig_app:keras_helper}.

\paragraph{Auxiliary functions}
\label{app:auxiliary_functions}

Figure~\ref{fig:lstm_cell} shows how we often design search spaces.
We have a high-level inductive bias (e.g., what operations are likely to be useful) for a good architecture for a task, but we might be unsure about low-level details (e.g., the exact sequence of operations of the architecture).
Auxiliary functions allows us to encapsulate aspects of search space creation and can be reused for creating different search spaces, e.g., through different calls to these functions.

\begin{figure}[htbp]
    \centering
\begin{minipage}{0.49\textwidth}
    \begin{align*}
        i_t &= \sigma(W_{ii} x_t + b_{ii} + W_{hi} h_{t-1} + b_{hi}) \\
        f_t &= \sigma(W_{if} x_t + b_{if} + W_{hf} h_{t-1} + b_{hf}) \\
        g_t &= \tanh(W_{ig} x_t + b_{ig} + W_{hg} h_{t-1} + b_{hg}) \\
        o_t &= \sigma(W_{io} x_t + b_{io} + W_{ho} h_{t-1} + b_{ho}) \\
        c_t &= f_t c_{t-1} + i_t g_t \\
        h_t &= o_t \tanh(c_t) \\ 
    \end{align*}
    \begin{align*}
        i_t &= q_i(x_t, h_{t-1}) \\
        f_t &= q_f(x_t, h_{t-1}) \\
        g_t &= q_g(x_t, h_{t-1}) \\
        o_t &= q_o(x_t, h_{t-1}) \\
        c_t &= q_c(f_t, c_{t-1}, i_t, g_t) \\
        h_t &= q_h(o_t, c_t)
    \end{align*}
\end{minipage} \hspace{0.1cm}
\begin{minipage}{0.48\textwidth}
\begin{lstlisting}[style = Python]
def lstm_cell(input_fn, forget_fn, gate_fn,
              output_fn, cell_fn, hidden_fn):

    x_inputs, x_outputs = identity()
    hprev_inputs, hprev_outputs = identity()
    cprev_inputs, cprev_outputs = identity()

    i_inputs, i_outputs = input_fn()
    f_inputs, f_outputs = forget_fn()
    g_inputs, g_outputs = gate_fn()
    o_inputs, o_outputs = output_fn()
    c_inputs, c_outputs = cell_fn()
    h_inputs, h_outputs = hidden_fn()

    i_inputs["in0"].connect(x_outputs["out"])
    i_inputs["in1"].connect(hprev_outputs["out"])
    f_inputs["in0"].connect(x_outputs["out"])
    f_inputs["in1"].connect(hprev_outputs["out"])
    g_inputs["in0"].connect(x_outputs["out"])
    g_inputs["in1"].connect(hprev_outputs["out"])
    c_inputs["in0"].connect(f_outputs["out"])
    c_inputs["in1"].connect(cprev_outputs["out"])
    c_inputs["in2"].connect(i_outputs["out"])
    c_inputs["in3"].connect(g_outputs["out"])
    o_inputs["in0"].connect(x_inputs["in"])
    o_inputs["in1"].connect(hprev_inputs["in"])
    h_inputs["in0"].connect(o_outputs["out"])
    h_inputs["in1"].connect(c_outputs["out"])

    return ({"x": x_inputs["in"],
             "hprev": hprev_inputs["in"],
             "cprev": cprev_inputs["in"]},
            {"c": c_outputs["out"],
             "h": h_outputs["out"]})
\end{lstlisting}
\end{minipage}
\caption{
    \textit{Left:} LSTM equations showing how the expert might abstract the LSTM structure into a general functional dependency.
    \textit{Right:} Auxiliary function for a LSTM cell that takes functions that return the search spaces for input, output, and forget gates, and the cell update, hidden state output, and context mechanisms and arranges them together to create the larger LSTM-like search space.}
\label{fig:lstm_cell}
\end{figure}

\section{Search space example}
\label{app:search_space_example}

 Figure~\ref{fig:enas_search_space} shows the recurrent cell search space introduced in~\cite{pham2018efficient} encoded in our language implementation.
 This search space is composed of a sequence of nodes.
 For each node, we choose its type and from which node output will it get its input.
 The cell output is the average of the outputs of all nodes after the first one.
 The encoding of this search space exemplifies the expressiveness of substitution modules.
 The cell connection structure is created through a substitution module that has hyperparameters representing where each node will get its input from.
 The substitution function that creates this cell takes functions that return inputs and outputs of the subsearch spaces for the input and intermediate nodes.
 Each subsearch space determines the operation performed by the node.
 While more complex than the other examples that we have presented, the same language constructs allow us to approach the encoding of this search space.
 Functions \code{cell}, \code{input\_node}, \code{intermediate\_node}, and \code{search\_space} define search spaces that are fully encapsulated and that therefore, can be reused for creating new search spaces.

\begin{figure}[htbp]
\centering
\begin{minipage}{\textwidth}
\begin{minipage}{0.47\textwidth}
\begin{lstlisting}[style = Python]
def cell(num_nodes,
         h_units,
         input_node_fn,
         intermediate_node_fn,
         combine_fn):

    def substitution_fn(dh):
        input_node = input_node_fn(h_units)
        inter_nodes = [
            intermediate_node_fn(h_units)
            for _ in range(1, num_nodes)
        ]
        nodes = [input_node] + inter_nodes

        for i in range(1, num_nodes):
            nodes[i][0]["in"].connect(
                nodes[dh[str(i)]][1]["out"])

        used_ids = set(dh.values())
        unused_ids = set(range(num_nodes)
            ).difference(used_ids)
        c_inputs, c_outputs = combine_fn(
                len(unused_ids))
        for j, i in enumerate(sorted(unused_ids)):
            c_inputs ["in%d"%j].connect(
                nodes[i][1]["out"])

        return (input_node[0],
                {"ht+1": c_outputs["out"]})

    name_to_hyperp = {str(i): D(range(i))
        for i in range(1, num_nodes)}

    return substitution_module("Cell",
        substitution_fn, name_to_hyperp,
        ["x", "ht"], ["ht+1"])
\end{lstlisting}
\end{minipage}
\hspace{0.55cm}
\begin{minipage}{0.47\textwidth}
\begin{lstlisting}[style = Python]
def input_node_fn(h_units):
    h_inputs, h_outputs = affine(h_units)
    x_inputs, x_outputs = affine(h_units)
    a_inputs, a_outputs = add(2)
    n_inputs, n_outputs = nonlinearity(D(["relu",
        "tanh","sigmoid", "identity"]))

    a_inputs["in0"].connect(x_outputs["out"])
    a_inputs["in1"].connect(h_outputs["out"])
    n_inputs["in"].connect(a_outputs["out"])

    return {
        "x": x_inputs["in"],
        "ht": h_inputs["in"]}, n_outputs


def intermediate_node_fn(h_units):
    a_inputs, a_outputs = affine(h_units)
    n_inputs, n_outputs = nonlinearity(D(["relu",
        "tanh", "sigmoid", "identity"]))
    a_outputs["out"].connect(n_inputs["in"])
    return a_inputs, n_outputs
\end{lstlisting}
\begin{lstlisting}[style = Python]
def search_space():
    h_units = D([32, 64, 128, 256])
    return cell(8, h_units,
        input_node_fn, intermediate_node_fn, avg)
\end{lstlisting}
\end{minipage}
\end{minipage}
\caption{Recurrent search space from ENAS~\cite{pham2018efficient} encoded using our language implementation.
A substitution module is used to delay the creation of the cell topology.
The code uses higher order functions to create the cell search space from the subsearch spaces of its nodes (i.e., \code{input\_node\_fn} and \code{intermediate\_node\_fn}).}
\label{fig:enas_search_space}
\end{figure}

\section{Additional details about language mechanics}
\label{app:mechanics_details}

\paragraph{Ordered module traversal}
Algorithm~\ref{alg:module_traversal} generates a unique ordering over modules $M(G)$ by starting at the modules that have outputs in $O_u(G)$ (which are named by $\sigma_o$) and traversing backwards, moving from a module to its neighboring modules (i.e., the modules that connect an output to an input of this module).
A unique ordering is generated by relying on the lexicographic ordering of the local names (see lines 3 and 10 in Algorithm~\ref{alg:module_traversal}).

\paragraph{Architecture instantiation}
Mapping an architecture $G \in \mathcal T$ relies on traversing $M(G)$ in topological order.
Intuitively, to do the local computation of a module $m \in M(G)$ for $G \in \mathcal T$, the modules that $m$ depends on (i.e., which feed an output into an input of $m$) must have done their local computations to produce their outputs (which will now be available as inputs to $m$).
Graph propagation (Algorithm~\ref{alg:graph_propagation}) starts with values for the unconnected inputs $I_u(G)$ and applies local module computation according to the topological ordering of the modules until the values for the unconnected outputs $O_u(G)$ are generated.
$g_{(m)}$ maps input and hyperparameter values to the local computation of $m$.
The arguments of $g_{(m)}$ and its results are sorted according to their local names (see lines 2 to 8).

\begin{figure}[htbp]
        \centering
\begin{minipage}{0.45\textwidth}
    \begin{algorithm}[H]
        \small
    \KwIn{$G \in \mathcal T, x_{(i)}$ for $i \in I_u(G)$ and $x_{(i)} \in \mathcal X_{(i)}$}
    \For{$m \in  \text{OrderedTopologically}(M(G))$}{
        $S_{(m), h} = \{s_{h, 1}, \ldots, s_{h, n_h}\}$ for $s_{h, 1} < \ldots < s_{h, n_h}$ \\
        $S_{(m), i} = \{s_{i, 1}, \ldots, s_{i, n_i}\}$ for $s_{i, 1} < \ldots < s_{i, n_i}$ \\
        $S_{(m), o} = \{s_{o, 1}, \ldots, s_{o, n_o}\}$ for $s_{o, 1} < \ldots < s_{o, n_o}$ \\
        $x_j \leftarrow x_{(\sigma_{(m), i}(s_{i, j}))}$, for $j \in [n_i]$ \\
        $v_j \leftarrow v_{(\sigma_{(m), h}(s_{h, j}))}$, for $j \in [n_h]$ \\
        $(y_1, \ldots, y_{n_o}) \leftarrow g_{(m)}(x_1, \ldots, x_{n_i}, v_1, \ldots, v_{n_h})$\\
        $y_{\sigma_{(m), o}(s_{o, j})} \leftarrow y_j$ for $j \in [n_o]$\\
        \For{$(o, i) \in E_o(m)$}{
            $x_{(i)} \leftarrow y_{(o)}$
        }
    }
    \Return{$y_{(o)}$ for $o \in O_u(G)$}
    \caption{Forward}
    \label{alg:graph_propagation}
    \end{algorithm}
\end{minipage}
 \hfill
        \begin{minipage}{0.48\textwidth}
            \centering
            \begin{algorithm}[H]
                \small
                \KwIn{$G, \sigma_o : S_o \to O_u(G)$}
                $M_q \leftarrow [\,]$ \\
                $n \leftarrow |S_o|$ \\
                Let $S_o = \{s_1, \ldots, s_{ n } \}$ with $s_1 < \ldots < s_{n}$.\\
                \For{$k \in [n]$}{
                    $m \leftarrow m(\sigma_o(s_k))$ \\
                    \If{$m \notin M_q$}{
                        $M_q \leftarrow M_q + [m]$
                    }
                }
                \For{$m \in M_q$}{
                    $n \leftarrow |S_{(m), i}|$ \\
                    Let $S_{(m), i} = \{s_1, \ldots, s_{ n } \}$ with $s_1 < \ldots < s_{n}$.\\
                    \While{$j \in [n]$}{
                        $i \leftarrow \sigma_{(m), i}(s_j)$ \\
                        \If{$i \notin I_u(G)$}{
                            Take $(o, i) \in E(G)$\\
                            $m' \leftarrow m(o)$ \\
                            \If{$m' \notin M_q$}{
                                $M_q \leftarrow M_q + [m']$ \\
                            }
                        }
                    }
                }
                \Return{$M_q$}
            \caption{OrderedModules}
            \label{alg:module_traversal}
            \end{algorithm}
        \end{minipage}
    \caption{
    \textit{Left:} \code{Forward} maps a terminal search space to its domain implementation.
    The mapping relies on each basic module doing its local computation (encapsulated by $g_{(m)}$ on line 7).
    \code{Forward} starts with values for the unconnected inputs and traverses the modules in topological order to generate values for the unconnected outputs.
    \textit{Right:} Iteration of $M(G)$ according to a unique order.
    The first \texttt{while} (line 4) loop adds the modules of the outputs in $O_u(G)$.
    The second \texttt{while} (line 8) loop traverses backwards the connections of the modules in $M_q$, adding new modules reached this way to $M_q$.
    $m(o)$ denotes the module that $o$ belongs to.
    See also Figure~\ref{fig:main_algorithms}
    }
\end{figure}

\section{Discussion about language expressivity}
\label{sec:expressivity}

\subsection{Infinite search spaces}
\label{sec:infinite_search_spaces}

\begin{wrapfigure}{r}{0.5\textwidth}
    \begin{lstlisting}[style = Python]
def maybe_one_more(fn):
    return siso_or([
            fn, lambda: siso_sequential(
                    [fn(), maybe_one_more(fn)])],
            D([0, 1]))
    \end{lstlisting}
    \caption{Self-similar search space either returns a search space or a search space and an optional additional search space.
    \code{fn} returns the search space to use in this construction.}
    \label{fig_app:infinite_search_space}
\end{wrapfigure}

We can rely on the laziness of substitution modules to encode infinite search spaces.
Figure~\ref{fig_app:infinite_search_space} shows an example of such a search space.
If the hyperparameter associated to the substitution module is assigned the value one, a new substitution module and hyperparameter are created.
If the hyperparameter associated to the substitution module is assigned the value zero, recursion stops.
The search space is infinite because the recursion can continue indefinitely.
This search space can be used to create other search spaces compositionally.
The same principles are valid for more complex search spaces involving recursion.

\subsection{Search space transformation and combination}
\label{ssec:hyperopt}

\begin{figure}[htbp]
    \centering
    \begin{minipage}{\textwidth}
    \begin{minipage}{0.47\textwidth}
\begin{lstlisting}[style = Python]
def search_space_1():
    return siso_repeat(
        lambda: siso_or([
            lambda: a_fn(D([0, 1])),
            lambda: b_fn(D([0, 1])),
            lambda: c_fn(D([0, 1]))],
        D([0, 1, 2])), D([1, 2, 4]))
\end{lstlisting}
\begin{lstlisting}[style = Python]
def search_space_2():
    return siso_or([
        lambda: siso_repeat(
            lambda: a_fn(D([0, 1])),
                D([1, 2, 4])),
        lambda: siso_repeat(
            lambda: b_fn(D([0, 1])),
                D([1, 2, 4])),
        lambda: siso_repeat(
            lambda: c_fn(D([0, 1])),
                D([1, 2, 4]))],
        D([0, 1, 2]))
\end{lstlisting}
    \end{minipage}
    \hspace{0.55cm}
\begin{minipage}{0.47\textwidth}
\begin{lstlisting}[style = Python]
def search_space_3():
    h = D([0, 1])
    return siso_or([
            lambda: siso_repeat(
                lambda: a_fn(h), D([1, 2, 4])),
            lambda: siso_repeat(
                lambda: b_fn(h), D([1, 2, 4])),
            lambda: siso_repeat(
                lambda: c_fn(h), D([1, 2, 4]))],
        D([0, 1, 2]))
\end{lstlisting}
\begin{lstlisting}[style = Python]
def search_space_4():
    return siso_or([
            lambda: siso_repeat(
                search_space_1, D([1, 2, 4])),
            lambda: siso_repeat(
                search_space_2, D([1, 2, 4])),
            lambda: siso_repeat(
                search_space_3, D([1, 2, 4]))],
        D([0, 1, 2]))
\end{lstlisting}
    \end{minipage}
    \end{minipage}
    \caption{
        \emph{Top left:} Repeats the choice between \code{a\_fn}, \code{b\_fn}, and \code{c\_fn} one, two, or four times.
        This search space shows that expressive search spaces can be created through simple arrangements of substitution modules.
        \emph{Bottom left:} Simple transformation of \code{search\_space\_1}.
        \emph{Top right:} Similar to \code{search\_space\_2}, but with the binary hyperparameter shared across all repetitions.
        \emph{Bottom right:} Simple search space that is created by composing the previously defined search spaces to create a new substitution module.
        }
\label{fig_app:search_space_transformations}
\end{figure}

We assume the existence of functions \code{a\_fn}, \code{b\_fn}, and \code{c\_fn} that each take one binary hyperparameter and return a search space.
In Figure~\ref{fig_app:search_space_transformations}, \code{search\_space\_1} repeats a choice between \code{a\_fn}, \code{b\_fn}, and \code{c\_fn} one, two, or four times.
The hyperparameters for the choice (i.e., those associated to \code{siso\_or}) modules are assigned values separately for each repetition.
The hyperparameters associated to each \code{a\_fn}, \code{b\_fn}, or \code{c\_fn} are also assigned values separately.

Simple rearrangements lead to dramatically different search spaces.
For example, we get \code{search\_space\_2} by swapping the nesting order of \code{siso\_repeat} and \code{siso\_or}.
This search space chooses between a repetition of one, two, or four \code{a\_fn}, \code{b\_fn}, or \code{c\_fn}.
Each binary hyperparameter of the repetitions is chosen separately.
\code{search\_space\_3} shows that it is simple to share an hyperparameter across the repetitions by instantiating it outside the function (line 2), and access it on the function (lines 5, 7, and 9).
\code{search\_space\_1}, \code{search\_space\_2}, and \code{search\_space\_3} are encapsulated and can be used as any other search space.
\code{search\_space\_4} shows that we can easily use \code{search\_space\_1}, \code{search\_space\_2}, and \code{search\_space\_3} in a new search space (compare to \code{search\_space\_2}).

Highly-conditional search spaces can be created through local composition of modules, reducing cognitive load.
In our language, substitution modules, basic modules, dependent hyperparameters, and independent hyperparameters are well-defined constructs to encode complex search spaces.
For example, \code{a\_fn} might be complex, creating many modules and hyperparameters, but its definition encapsulates all this.
This is one of the greatest advantages of our language, allowing us to easily create new search spaces from existing search spaces.
Furthermore, the mapping from instances in the search space to implementations is automatically generated from the search space encoding.

\section{Implementation details}
\label{sec:implementation_details}

This section gives concrete details about our Python language implementation.
We refer the reader to \url{https://github.com/negrinho/deep_architect} for additional code and documentation.

\subsection{Supporting new domains}
\label{sec:domain_support}

We only need to extend \code{Module} class to support basic modules in the new domain.
We start with the common implementation of \code{Module} (see Figure~\ref{fig:module_implementation}) for both basic and substitution modules and then cover its extension to support Keras basic modules (see Figure~\ref{fig_app:keras_helper}).

\begin{figure}[tbp]
    \centering
\begin{lstlisting}[style = Python]
class Module(Addressable):

    def __init__(self, scope=None, name=None):
        scope = scope if scope is not None else Scope.default_scope
        name = scope.get_unused_name('.'.join(
            ['M', (name if name is not None else self._get_base_name()) + '-']))
        Addressable.__init__(self, scope, name)

        self.inputs = OrderedDict()
        self.outputs = OrderedDict()
        self.hyperps = OrderedDict()
        self._is_compiled = False

    def _register_input(self, name):
        assert name not in self.inputs
        self.inputs[name] = Input(self, self.scope, name)

    def _register_output(self, name):
        assert name not in self.outputs
        self.outputs[name] = Output(self, self.scope, name)

    def _register_hyperparameter(self, name, h):
        assert isinstance(h, Hyperparameter) and name not in self.hyperps
        self.hyperps[name] = h
        h._register_module(self)

    def _register(self, input_names, output_names, name_to_hyperp):
        for name in input_names:
            self._register_input(name)
        for name in output_names:
            self._register_output(name)
        for name in sorted(name_to_hyperp):
            self._register_hyperparameter(name, name_to_hyperp[name])

    def _get_input_values(self):
        return {name: ix.val for name, ix in iteritems(self.inputs)}

    def _get_hyperp_values(self):
        return {name: h.get_value() for name, h in iteritems(self.hyperps)}

    def _set_output_values(self, output_name_to_val):
        for name, val in iteritems(output_name_to_val):
            self.outputs[name].val = val

    def get_io(self):
        return self.inputs, self.outputs

    def get_hyperps(self):
        return self.hyperps

    def _update(self):
        """Called when an hyperparameter that the module depends on is set."""
        raise NotImplementedError

    def _compile(self):
        raise NotImplementedError

    def _forward(self):
        raise NotImplementedError

    def forward(self):
        if not self._is_compiled:
            self._compile()
            self._is_compiled = True
        self._forward()
\end{lstlisting}
\caption{
    \code{Module} class used to implement both basic and substitution modules.
    \code{\_register\_input}, \code{\_register\_output}, \code{\_register\_hyperparameter}, \code{\_register}, \code{\_get\_hyperp\_values}, \code{get\_io} and \code{get\_hyperps} are used by both basic and substitution modules.
    \code{\_get\_input\_values}, \code{\_set\_output\_values}, \code{\_compile}, \code{\_forward}, and \code{forward} are used only by basic modules.
    \code{\_update} is used only by substitution modules. }
\label{fig:module_implementation}
\end{figure}

\paragraph{General module class}
The complete implementation of \code{Module} is shown in Figure~\ref{fig:module_implementation}.
\code{Module} supports the implementations of both basic modules and substitution modules.
There are three types of functions in \code{Module} in Figure~\ref{fig:module_implementation}: those that are used by both basic and substitution modules (\code{\_register\_input}, \code{\_register\_output}, \code{\_register\_hyperparameter}, \code{\_register}, \code{\_get\_hyperp\_values}, \code{get\_io} and \code{get\_hyperps}); those that are used just by basic modules (\code{\_get\_input\_values}, \code{\_set\_output\_values}, \code{\_compile}, \code{\_forward}, and \code{forward}); those are used just by substitution modules (\code{\_update}).
We will mainly discuss its extension for basic modules as substitution modules are domain-independent (e.g., there are no domain-specific components in the substitution modules in Figure~\ref{fig_app:substitution_modules} and in \code{cell} in Figure~\ref{fig:enas_search_space}).

Supporting basic modules in a domain relies on two functions: \code{\_compile} and \code{\_forward}.
These functions help us map an architecture to its implementation in deep learning (slightly different functions might be necessary for other domains).
\code{forward} shows how \code{\_compile} and \code{\_forward} are used during graph instantiation in a terminal search space.
See Figure~\ref{fig:forward} for the iteration over the graph in topological ordering (determined by \code{determine\_module\_eval\_seq}), and evaluates the forward calls in turn for the modules in the graph leading to its unconnected outputs.

\code{\_register\_input}, \code{\_register\_output}, \code{\_register\_hyperparameter}, and \code{\_register} are used to describe the inputs and outputs of the module (i.e., \_register\_input and \_register\_output), and to associate hyperparameters to its properties (i.e., \code{\_register\_hyperparameter}).
\code{\_register} aggregates the first three functions into one.
\code{\_get\_hyperp\_values}, \code{\_get\_input\_values}, and \code{\_set\_output\_values} are used in \code{\_forward} (see left of Figure~\ref{fig_app:keras_helper}.
These are used in each basic module, once in a terminal search space, to retrieve its hyperparameter values (\code{\_get\_hyperp\_values}) and its input values (\code{\_get\_input\_values}) and to write the results of its local computation to its outputs (\code{\_set\_output\_values}).
Finally, \code{get\_io} retrieves the dictionaries mapping names to inputs and outputs (these correspond to $\sigma_{(m), i} : S_{(m), i} \to I(m)$ and $\sigma_{(m), o} : S_{(m), o} \to O(m)$, respectively, described in Section~\ref{sec:implementation_of_language}).
Most inputs are named \code{in} if there is a single input and \code{in0}, \code{in1}, and so on if there is more than one.
Similarly, for outputs, we have \code{out} for a single output, and \code{out0}, \code{out1}, and so if there are multiple outputs.
This is often seen when connecting search spaces, e.g., lines 15 to 28 in right of Figure~\ref{fig:lstm_cell}.
In Figure~\ref{fig:lstm_cell}, we redefine $\sigma_i$ and $\sigma_o$ (in line 30 to line 34) to have appropriate names for the LSTM cell, but often, if possible, we just use $\sigma_{(m), i}$ and $\sigma_{(m'), o}$ for $\sigma_i$ and $\sigma_o$ respectively, e.g., in \code{siso\_repeat} and \code{siso\_combine} in Figure~\ref{fig_app:substitution_modules}.

\code{\_update} is used in substitution modules (not shown in Figure~\ref{fig:module_implementation}): for a substitution module, it checks if all its hyperparameters have been assigned values and does the substitution (i.e., calls its substitution function to create a search space that takes the place of the substitution module; e.g., see frames \code{a}, \code{b}, and \code{c} of Figure~\ref{fig:graph_transitions} for a pictorial representation, and Figure~\ref{fig_app:substitution_modules} for implementations of substitution modules).
In the examples of Figure~\ref{fig_app:substitution_modules}, \code{substitution\_fn} returns the search space to replace the substitution module with in the form of a dictionary of inputs and a dictionary of outputs (corresponding to $\sigma_i$ and $\sigma_o$ on line 12 of Algorithm~\ref{alg:graph_transition}).
The substitution modules that we considered can be implemented with the helper in Figure~\ref{fig_app:substitution_module} (e.g., see the examples in Figure~\ref{fig_app:substitution_modules}).

In the signature of \code{\_\_init\_\_} for \code{Module}, \code{scope} is a namespace used to register a module with a unique name and \code{name} is the prefix used to generate the unique name.
Hyperparameters also have a unique name generated in the same way.
Figure~\ref{fig:graph_transitions} shows this in how the modules and hyperparameters are named, e.g., in frame \code{a}, \code{Conv2D-1} results from generating a unique identifier for \code{name} \code{Conv2D} (this is also captured in the use of \code{\_get\_name} in the examples in Figure~\ref{fig_app:basic_modules} and Figure~\ref{fig_app:substitution_modules}).
When \code{scope} is not mentioned explicitly, a default global scope is used (e.g., \code{scope} is optional in Figure~\ref{fig:module_implementation}).

\paragraph{Extending the module class for a domain (e.g., Keras)}
Figure~\ref{fig_app:keras_helper} (left) shows the extension of \code{Module} to deal with basic modules in Keras.
\code{KerasModule} is the extension of \code{Module}.
\code{keras\_module} is a convenience function that instantiates a \code{KerasModule} and returns its dictionary of local names to inputs and outputs.
\code{siso\_keras\_module} is the same as \code{keras\_module} but uses default names \code{in} and \code{out} for a single-input single-output module, which saves the expert the trouble of explicitly naming inputs and outputs for this common case.
Finally, \code{siso\_keras\_module\_from\_keras\_layer\_fn} reduces the effort of creating basic modules from Keras functions (i.e., the function can be passed directly creating \code{compile\_fn} beforehand).
These functions are analogous for different deep learning frameworks, e.g., see the example usage of \code{siso\_tensorflow\_module} in Figure~\ref{fig_app:basic_modules}.

The most general helper, \code{keras\_module} works by providing the local names for the inputs (\code{input\_names}) and outputs (\code{output\_names}), the dictionary mapping local names to hyperparameters (\code{name\_to\_hyperp}), and the compilation function (\code{compile\_fn}), which corresponds to the \code{\_compile\_fn} function of the module.
Calling \code{\_compile\_fn} returns a function (corresponding to \code{\_forward} for a module, e.g., see Figure~\ref{fig_app:basic_modules}).

\begin{figure}[htbp]
    \centering
    \begin{minipage}{\textwidth}
    \begin{minipage}{0.55\textwidth}
    \begin{lstlisting}[style = Python]
import deep_architect.core as co

class KerasModule(co.Module):

    def __init__(self,
                    name,
                    compile_fn,
                    name_to_hyperp,
                    input_names,
                    output_names,
                    scope=None):
        co.Module.__init__(self, scope, name)
        self._register(input_names, output_names,
            name_to_hyperp)
        self._compile_fn = compile_fn

    def _compile(self):
        input_name_to_val = self._get_input_values()
        hyperp_name_to_val = self._get_hyperp_values()
        self._fn = self._compile_fn(
            input_name_to_val, hyperp_name_to_val)

    def _forward(self):
        input_name_to_val = self._get_input_values()
        output_name_to_val = self._fn(input_name_to_val)
        self._set_output_values(output_name_to_val)

    def _update(self):
        pass
\end{lstlisting}
    \end{minipage}
    \hspace{0.55cm}
\begin{minipage}{0.42\textwidth}
\begin{lstlisting}[style = Python]
def keras_module(name,
                 compile_fn,
                 name_to_hyperp,
                 input_names,
                 output_names,
                 scope=None):
    return KerasModule(name, compile_fn,
        name_to_hyperp, input_names,
        output_names, scope).get_io()


def siso_keras_module(name, compile_fn,
        name_to_hyperp, scope=None):
    return KerasModule(name, compile_fn,
        name_to_hyperp, ['in'], ['out'],
                        scope).get_io()


def siso_keras_module_from_keras_layer_fn(
        layer_fn, name_to_hyperp,
        scope=None, name=None):

    def compile_fn(di, dh):
        m = layer_fn(**dh)

        def forward_fn(di):
            return {"out": m(di["in"])}

        return forward_fn

    if name is None:
        name = layer_fn.__name__

    return siso_keras_module(name,
        compile_fn, name_to_hyperp, scope)
\end{lstlisting}
    \end{minipage}
    \end{minipage}
    \caption{
        \emph{Left:} Complete extension of the \code{Module} class (see Figure~\ref{fig:module_implementation} for supporting Keras basic modules.
        \emph{Right:} Convenience functions to reduce the effort of wrapping Keras operations into basic modules for common cases.
        See Figure~\ref{fig_app:basic_modules} for examples of how they are used.
        }
    \label{fig_app:keras_helper}
    \end{figure}

\begin{figure}[tbp]
    \centering
\begin{lstlisting}[style = Python]
def forward(input_to_val, _module_seq=None):
    if _module_seq is None:
        _module_seq = determine_module_eval_seq(input_to_val.keys())

    for ix, val in iteritems(input_to_val):
        ix.val = val

    for m in _module_seq:
        m.forward()
        for ox in itervalues(m.outputs):
            for ix in ox.get_connected_inputs():
                ix.val = ox.val
\end{lstlisting}
\caption{
    Generating the implementation of the architecture in a terminal search space $G$ (e.g., the one in frame \code{d} of Figure~\ref{fig:graph_transitions}).
    Compare to Algorithm~\ref{alg:graph_propagation}: \code{input\_to\_val} corresponds to the $x_{(i)}$ for $i \in I_u(G)$;
    \code{determine\_module\_eval\_seq} corresponds to \code{OrderedTopologically} in line 1 of Algorithm~\ref{alg:graph_propagation};
    Remaining code corresponds to the traversal of the modules according to this ordering, evaluation of their local computations, and propagation of results from outputs to inputs.
}
\label{fig:forward}
\end{figure}

\subsection{Implementing a search algorithm}
\label{sec:search_algorithm_implementation}

\begin{figure}[tbp]
\begin{lstlisting}[style = Python]
def random_specify_hyperparameter(hyperp):
    assert not hyperp.has_value_assigned()

    if isinstance(hyperp, hp.Discrete):
        v = hyperp.vs[np.random.randint(len(hyperp.vs))]
        hyperp.assign_value(v)
    else:
        raise ValueError
    return v

def random_specify(outputs):
    hyperp_value_lst = []
    for h in co.unassigned_independent_hyperparameter_iterator(outputs):
        v = random_specify_hyperparameter(h)
        hyperp_value_lst.append(v)
    return hyperp_value_lst

class RandomSearcher(Searcher):
    def __init__(self, search_space_fn):
        Searcher.__init__(self, search_space_fn)

    def sample(self):
        inputs, outputs = self.search_space_fn()
        vs = random_specify(outputs)
        return inputs, outputs, vs, {}

    def update(self, val, searcher_eval_token):
        pass

\end{lstlisting}
\caption{Implementation of random search in our language implementation.
\code{sample} assigns values to all the independent hyperparameters in the search space, leading to an architecture that can be evaluated.
\code{update} incorporates the results of evaluating an architecture into the state of the searcher, allowing it to use this information in the next call to \code{sample}.
}
\label{fig:random_search_implementation}
\end{figure}

Figure~\ref{fig:random_search_implementation} shows random search in our implementation.
\code{random\_specify\_hyperparameter} assigns a value uniformly at random to an independent hyperparameter.
\code{random\_specify} assigns all unassigned independent hyperparameters in the search space until reaching a terminal search space (each assignment leads to a search space transition; see Figure~\ref{fig:graph_transitions}).
\code{RandomSearcher} encapsulates the behavior of the searcher through two main functions: \code{sample} and \code{update}.
\code{sample} samples an architecture from the search space, which returns \code{inputs} and \code{outputs} for the sampled terminal search space, the sequence of value assignments that led to the sampled terminal search space, and a \code{searcher\_eval\_token} that allows the searcher to identify the sampled terminal search space when the evaluation results are passed back to the searcher through a call to \code{update}.
\code{update} incorporates the evaluation results (e.g., validation accuracy) of a sampled architecture into the state of the searcher, allowing it to use this information in the next call to \code{sample}.
For random search, \code{update} is a no-op.
\code{\_\_init\_\_} takes the function returning a search space (e.g., \code{search\_space} in Figure~\ref{fig:enas_search_space}) from which architectures are to be drawn from and any other arguments that the searcher may need (e.g., exploration term in MCTS).
To implement a new searcher, \code{Searcher} needs to be extended by implementing \code{sample} and \code{update} for the desired search algorithm.
\code{unassigned\_independent\_hyperparameter\_iterator} provides ordered iteration over the independent hyperparameters of the search space.
The role of the search algorithm is to pick values for each of these hyperparameters, leading to a terminal space.
Compare to Algorithm~\ref{alg:random_search}.
\code{search\_space\_fn} returns the dictionaries of inputs and outputs for the initial state of the search space (analogous to the search space in frame \code{a} in Figure~\ref{fig:graph_transitions}).

\section{Additional experimental results}
\label{app:additional_results}

We present the full validation and test results for both the search space experiments (Table~\ref{tab:ss_results}) and the search algorithm experiments (Table~\ref{tab:se_results}).
For each search space, we performed a grid search over the learning rate with values in $\{0.1, 0.05, 0.025, 0.01, 0.005, 0.001\}$ and an L2 penalty with values in $\{0.0001, 0.0003, 0.0005\}$ for the architecture with the highest validation accuracy
Each evaluation in the grid search was trained for 600 epochs with SGD with momentum of $0.9$ and a cosine learning rate schedule
We did a similar grid search for each search algorithm.

\begin{table}[htbp]
  \centering
    \caption{
    Results for the search space experiments
    A grid search was performed on the best architecture from the search phase
    Each evaluation in the grid search was trained for 600 epochs
    }
  \begin{tabular}{lcccc}
    \toprule
    Search Space     & \breakcell{Validation Accuracy\\@ 25 epochs} & \breakcell{Validation Accuracy\\@ 600 epochs} & \breakcell{Test Accuracy\\@ 600 epochs} & \breakcell{Number of\\Parameters}\\
    \midrule
    Genetic~\cite{xie2017genetic}       & 79.03 & 91.13 & 90.07 & 9.4M \\
    Flat~\cite{liu2017hierarchical}     & 80.69 & 93.70 & 93.58 & 11.3M \\
    Nasbench~\cite{ying2019nasbench}    & 87.66 & 95.08 & 94.59 & 2.6M \\
    Nasnet~\cite{zoph2018learning}      & 82.35 & 94.56 & 93.77 & 4.5M \\
    \bottomrule
  \end{tabular}
  \label{tab:ss_results}
\end{table}

\begin{table}[htbp]
  \centering
    \caption{
    Results for the search algorithm experiments
    A grid search was performed on the best architecture from the search phase, each trained to 600 epochs
    }
  \begin{tabular}{lcccc}
    \toprule
    Search algorithm     & Run & \breakcell{Validation\\Accuracy\\@ 25 epochs} & \breakcell{Validation\\Accuracy\\@ 600 epochs} & \breakcell{Test\\Accuracy\\@ 600 epochs}\\
    \midrule
    \multirow{4}{*}{Random}
        & 1 & 77.58 & 92.61 & 92.38 \\
        &2 & 79.09 & 91.93 & 91.30 \\
        &3 & 81.26 & 92.35 & 91.16 \\
        &\textbf{Mean} & ${79.31 \pm 1.85}$ & ${92.29 \pm 0.34}$ & ${91.61 \pm 0.67}$ \\
        \hline
    \multirow{4}{*}{MCTS~\cite{browne2012survey}}
        & 1 & 78.68 & 91.97 & 91.33 \\
        &2 & 78.65 & 91.59 & 91.47 \\
        &3 & 78.65 & 92.69 & 91.55 \\
        &\textbf{Mean} & ${78.66 \pm 0.02}$ & ${92.08 \pm 0.56}$ & ${91.45 \pm 0.11}$ \\
        \hline
    \multirow{4}{*}{SMBO~\cite{negrinho2017deeparchitect}}
        & 1 & 77.93 & 93.62 & 92.92 \\
        &2 & 81.80 & 93.05 & 92.03 \\
        &3 & 82.73 & 91.89 & 90.86 \\
        &\textbf{Mean} & ${80.82 \pm 2.54}$ & ${92.85 \pm 0.88}$ & ${91.93 \pm 1.03}$ \\
    \hline
    \multirow{4}{*}{Regularized evolution~\cite{real2018regularized}}
        & 1 & 80.99 & 92.06 & 90.80 \\
        &2 & 81.51 & 92.49 & 91.79 \\
        &3 & 81.65 & 92.10 & 91.39 \\
        &\textbf{Mean} & ${81.38 \pm 0.35}$ & ${92.21 \pm 0.24}$ & ${91.32 \pm 0.50}$ \\
    \bottomrule
  \end{tabular}
  \label{tab:se_results}
\end{table}

\end{document}